\documentclass{egpubl}

\usepackage{eg2024}
 
\STAR                   
\CGFStandardLicense

\usepackage[T1]{fontenc}
\usepackage{dfadobe}  

\BibtexOrBiblatex
\usepackage[backend=biber,bibstyle=EG,citestyle=alphabetic,backref=true,mincitenames=1,maxcitenames=2,natbib=true]{biblatex} 
\renewcommand*{\mkbibnamefamily}[1]{#1}
\renewcommand*{\mkbibnamegiven}[1]{#1}
\electronicVersion
\PrintedOrElectronic
\ifpdf \usepackage[pdftex]{graphicx} \pdfcompresslevel=9
\else \usepackage[dvips]{graphicx} \fi

\usepackage{egweblnk} 

\usepackage{url}
\usepackage{amsmath,amsfonts,amssymb}
\usepackage{xspace}
\usepackage{hyperref}
\usepackage{nicefrac}

\usepackage{cleveref}
\usepackage[table]{xcolor}
\usepackage{booktabs}
\usepackage{makecell}
\usepackage{tabularx}
\usepackage{multirow}
\usepackage{colortbl}
\usepackage{titlesec} %

\definecolor{verylightgray}{gray}{0.95}

\DeclareMathOperator{\Enc}{Enc}
\DeclareMathOperator{\Dec}{Dec}
\newcommand{\Input}{\ensuremath{x}}
\newcommand{\InputVoxel}{\ensuremath{\Input^V}}
\newcommand{\InputImage}{\ensuremath{\Input^I}}
\newcommand{\InputText}{\ensuremath{\Input^T}}

\newcommand{\EncVoxel}{\ensuremath{\Enc^V}}
\newcommand{\DecVoxel}{\ensuremath{\Dec^V}}
\newcommand{\Emb}{\ensuremath{e}}
\newcommand{\EmbVoxel}{\ensuremath{\Emb^V}}

\newcommand{\CLIP}{\text{CLIP}}
\newcommand{\EncCLIPText}{\ensuremath{\Enc^T_{\CLIP}}}
\newcommand{\EncCLIPImage}{\ensuremath{\Enc^I_{\CLIP}}}
\newcommand{\EmbCLIPText}{\ensuremath{\Emb^T_{\CLIP}}}
\newcommand{\EmbCLIPImage}{\ensuremath{\Emb^I_{\CLIP}}}
\newcommand{\EmbCLIP}{\ensuremath{\Emb_{\CLIP}}}

\newcommand{\paired}{\textsc{3DPT}\xspace}
\newcommand{\unpaired}{\textsc{3DUT}\xspace}
\newcommand{\notd}{\textsc{No3D}\xspace}

\definecolor{lightblue}{RGB}{0,128,255}
\newcommand{\edited}[1]{\textcolor{black}{#1}}

\newcommand{\mypara}[1]{\vspace{2pt}\noindent\textbf{#1}}

\newcommand{\notavail}[1]{{\color{gray}{#1}}}

\usepackage{amsmath,amsfonts,bm}

\def\eqref#1{equation~\ref{#1}}

\def\1{\bm{1}}

\DeclareMathAlphabet{\mathsfit}{\encodingdefault}{\sfdefault}{m}{sl}
\SetMathAlphabet{\mathsfit}{bold}{\encodingdefault}{\sfdefault}{bx}{n}

\newcommand{\E}{\mathbb{E}}

\title{Text-to-3D Shape Generation}

\author[H. Lee, M. Savva \& A. X. Chang]{%
H. Lee$^{1}$ \quad M. Savva$^{1}$ \quad A. X. Chang$^{1,2}$\vspace{-10pt}\\
$^1$Simon Fraser University \quad $^2$Canada CIFAR AI Chair, Amii\vspace{4pt}\\
\href{https://3dlg-hcvc.github.io/tt3dstar/}{3dlg-hcvc.github.io/tt3dstar}%
}

\begin{document}

\teaser{
\includegraphics[trim=0 0 0 0,clip,width=\linewidth]{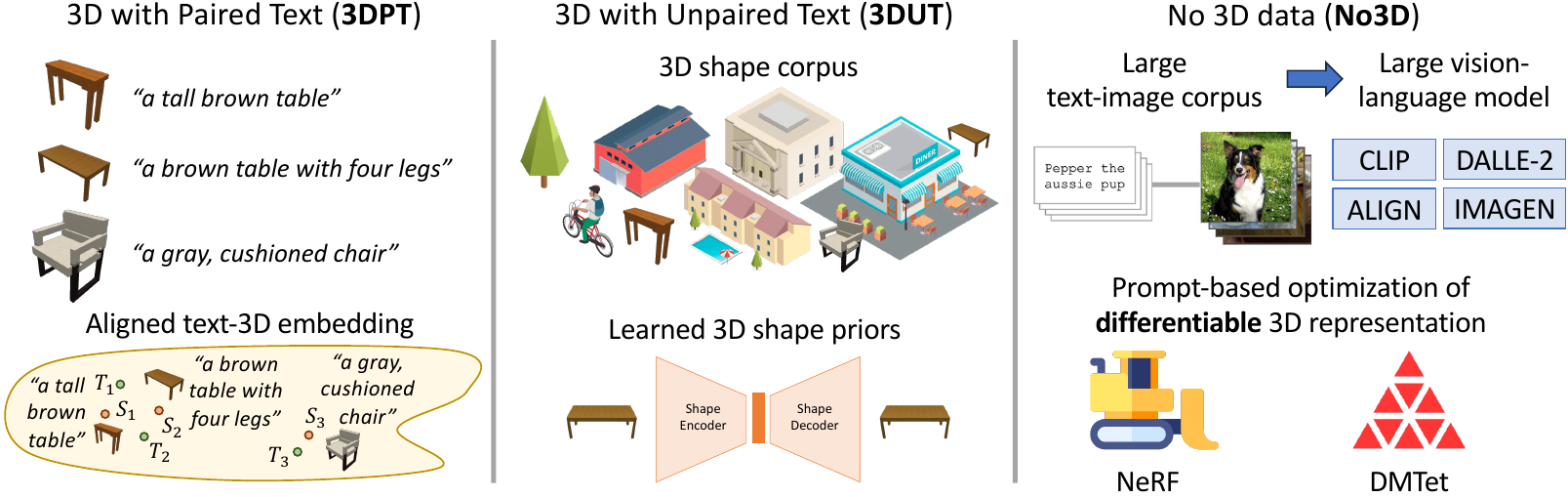} 
\centering
\caption{We survey and categorize methods for text-to-3D shape generation. Our categorization organizes methods into three families delineated by the use of 3D and text data. The first and second family rely on 3D data in combination with either paired text descriptions (\paired) or with no text-3D pairing (\unpaired).  The third family does not rely on 3D data for training (\notd).}
\label{fig:overview}
}

\maketitle
\begin{abstract}
Recent years have seen an explosion of work and interest in text-to-3D shape generation.  Much of the progress is driven by advances in 3D representations, large-scale pretraining and representation learning for text and image data enabling generative AI models, and differentiable rendering.  Computational systems that can perform text-to-3D shape generation have captivated the popular imagination as they enable non-expert users to easily create 3D content directly from text. However, there are still many limitations and challenges remaining in this problem space.  In this state-of-the-art report, we provide a survey of the underlying technology and methods enabling text-to-3D shape generation to summarize the background literature. We then derive a systematic categorization of recent work on text-to-3D shape generation based on the type of supervision data required.  Finally, we discuss limitations of the existing categories of methods, and delineate promising directions for future work.
\end{abstract}

\section{Introduction}

\begin{figure*}
\centering
\includegraphics[width=\textwidth]{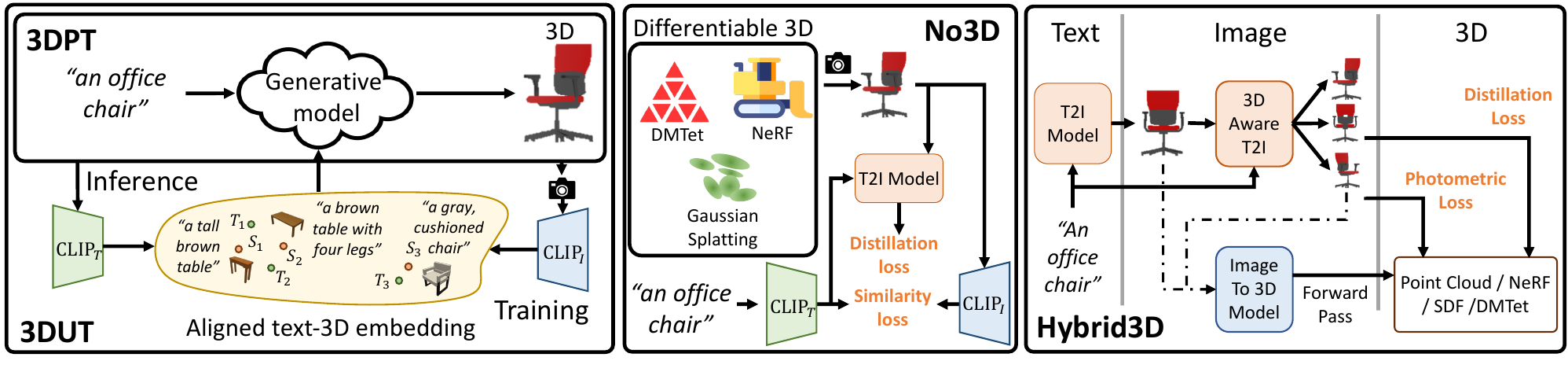}
\caption{\edited{We illustrate the main components for the different families of methods in this survey. 3DPT methods involve training a generative model on paired text and 3D data. In 3DUT, data is limited to only 3D shapes, so methods in this group leverage the aligned text-3D embedding space of CLIP encoders allowing conditioning on rendered images during training and text prompts during inference. In No3D, no data is available so methods rely on pretrained guidance models such as CLIP and T2I diffusion models to optimize similarity and distillation respectively with differentiable 3D representations. 
Finally, Hybrid3D methods often combine techniques from \notd with priors learned from 3D assets to train 3D-aware T2I models or to enable text-to-image-to-3D using pretrained large text-to-image models.
}}
\label{fig:overview-models}
\end{figure*}

Text to 3D shape generation methods can revolutionize 3D content creation by allowing anyone to generate 3D content based on a simple text description.
It is no wonder that there has been an explosion of interest in this research direction.
Recent advances in generative models for text and images~\cite{ramesh2022hierarchical,saharia2022photorealistic} enabled by large-scale language and vision-language models, as well as advances in learned 3D representations and 3D generative models have acted as catalysts for progress in text to 3D shape generation.

At the same time as this rapid progress, a number of open research challenges are coming into focus.
There is currently a sparsity of available 3D data paired with natural language text descriptions, making it infeasible to rely purely on direct supervision from data pairs in both domains.
Moreover, current text to 3D generation methods do not afford natural editability of the generated outputs in an intuitive way based on user inputs.
Generation of larger-scale 3D outputs representing compositions of objects into natural scenes also remains challenging.
Lastly, the complexity of underlying 3D generative models coupled with the complexity of the optimization problem when avoiding reliance on paired text and 3D data lead to a challenging learning problem with significant compute and training time requirements.

There have been some recent related surveys addressing 3D generative models~\cite{chaudhuri2020learning,shi2022deep} and text-conditioned generation~\cite{chao2023text,li2023generative}.
The former group of surveys on 3D generative models addresses what is typically one component of a complete text-to-3D shape system.
\citet{chao2023text} cover both text-to-image and text-to-3D generation and do not focus on a detailed categorization and discussion of 3D generation methods.
\citet{li2023generative} do focus on text-to-3D generation specifically but address mainly application domains and do not derive a comprehensive categorization of the methods themselves, or pursue a systematic discussion of the method design decisions.

In this survey, we systematically summarize work on text-to-3D shape generation on the basis of four key components: training data, 3D representation type, generative model, and training setup.
For the first, we further categorize by the amount and type of 3D data that is needed for supervision.
Specifically, we categorize methods based on whether they:
1) use 3D data to train a generative model (\Cref{sec:method-3ddata}); or
2) tackle text-to-3D without any 3D data (\Cref{sec:method-unsupervised}).
\Cref{fig:overview} and \Cref{tab:method-category-comparison} provide a summary of this categorization.
We note that overall techniques in 1) are similar to text-to-image methods, with the key differences being: a) the 3D generative model; and b) the link between text and 3D.
For methods in 1), we further breakdown between methods that are supervised with paired text-3D data (\Cref{sec:method-3ddata-paired}) and methods that do not rely on paired data (\Cref{sec:method-3ddata-unpaired}).   As other surveys have focused purely on 3D generative models, in this survey we focus particularly on methods for text-to-3D without any 3D data, and how the link between text and 3D can be established without paired text to 3D data (typically leveraging 2D image information).
In addition, we discuss some of the key challenges of generating high quality 3D shapes without explicit 3D training data, and strategies that are employed to tackle these challenges.
\edited{Recent work has started to address these challenges by learning shape priors from large 3D datasets, often coupled with techniques introduced for text-to-3D without 3D data.
We describe these `hybrid' approaches in \Cref{sec:hybrid3d}.}

\section{Scope of this Survey}

\subsection{Comparison with Related Surveys}

\citet{shi2022deep} offers a comprehensive review of generative models for 3D data. They primarily concentrate on various 3D representations, including voxels, point clouds, neural fields, and meshes, and delve into different generative model types like GANs, VAEs, and energy-based models used by various works. While some papers that use text for conditional 3D generation are mentioned, it is not the main focus of their survey.

An earlier survey \citet{chaudhuri2020learning} in comparison emphasizes structure-aware generative models for 3D shapes and scenes. They focus on approaches that decompose 3D objects or scenes into smaller elements and introduces overarching structures using representations like hierarchies and programs. They showcase various generative models applied to these structures, ranging from classical probabilistic models to deep learning techniques.

\citet{chao2023text} present a survey on text-conditioned editing, encompassing both 2D and 3D methodologies. Though the emphasis is on editing, the survey details the key text-to-3D generation techniques that underpin these editing methods. Notably, the study predominantly highlights works that leverage CLIP for shape generation and editing. However, recent advancements using diffusion distillation methods remain unaddressed.

A recent review by \citet{li2023generative} examines recent text-to-3D research, notably those incorporating CLIP and diffusion models. Although there is some overlap in the work discussed, our survey offers a more structured categorization for easier comparison and provides more in-depth examination of each work.

\begin{table}[t]
\resizebox{\linewidth}{!}
{
\begin{tabular}{@{} p{10cm} @{}}
\toprule
\textbf{3D Paired Text (\paired)}: Requires paired text-3D data which is limited. Generation limited to observed data. \\
\midrule
\textbf{3D Unpaired Text (\unpaired)}: Leverages 3D data to train 3D generative model.  Bridges text and 3D using images. Can use vision-language models to generate captions for 3D data, reducing to ``Paired'' scenario.\\
\midrule
\textbf{No 3D Data (\notd)}: No 3D data for training.  Multi-view and structure consistency is an issue.  Uses images as bridge, typically with differentiable rendering.  Conceptually can generate arbitrary 3D content.  Per-prompt optimization, slow.\\
\midrule
\textbf{Hybrid3D}: Combine text-to-image and image-to-3D methods. Enforce 3D consistency using 3D--aware text-to-image models or multi-view images.\\
\bottomrule
\end{tabular}
}
\caption{Properties of the four families into which we categorize text-to-3D shape generation methods.}
\label{tab:method-category-comparison}
\end{table}

\subsection{Organization of this Survey}

As summarized in \Cref{tab:method-category-comparison}, our survey classifies recent text-to-3D research into three primary families:
1) Methods that utilize paired text with 3D data (\paired), discussed in \Cref{sec:method-3ddata-paired};
2) Methods reliant on 3D data but not requiring paired 3D and text data (\unpaired), discussed in \Cref{sec:method-3ddata-unpaired};
3) Methods that require no 3D data training data at all, and typically use CLIP or Diffusion models as guidance (\notd), discussed in \Cref{sec:method-unsupervised}; and
4) Recent work that leverage a combination of text-to-image and image-to-3D, discussed in \Cref{sec:hybrid3d}.

The main focus of this survey is primarily on the third family of works as they have not been addressed in detail by prior surveys.
For these \notd approaches we further divide into sections that discuss methods that:
1) leverage pre-trained text-image embeddings (\Cref{sec:method-unsupervised-clip})
2) formulate or improve upon ways to use diffusion models as a prior (\Cref{sec:method-unsupervised-diffusion});
3) use different 3D representations, rendering techniques, or improvements to the training setup to enhance the quality of the results.
\edited{\Cref{fig:overview-models} shows an overview of different methods covered in this survey.}

We then discuss emerging work on generation of multi-object 3D scenes (\Cref{sec:method-scene}), and on allowing editing of the output 3D shape in various ways (\Cref{sec:method-editing}).
The following section of the survey presents a brief overview of evaluation methods for text-to-3D shape generation (\Cref{sec:evaluation}).
We conclude the survey with a discussion of promising future directions (\Cref{sec:discussion}).

\section{Preliminaries}

In this section, we provide a brief background of fundamentals used in text-to-3D generation methods.
In particular, we summarize choices for 3D representations (\Cref{sec:bg-3dreps}), the necessary background on deep generative models (\Cref{sec:bg-generative}), and guidance models for connecting 3D representations with text via images (\Cref{sec:bg-guidance}).

\subsection{3D Representations}
\label{sec:bg-3dreps}

There are a variety of representations that are possible when working with 3D data, each with specific properties and challenges.
In their survey of 3D generative models, \citet{shi2022deep} include a good overview of the different choices.
Here we provide a brief summary, with a focus on how appearance (color or texture) is encoded in 3D models as these attributes are commonly included in text descriptions.
We also describe in detail neural radiance fields (NeRFs) and DMTet, as these are two popular representations used in text-to-3D generation.

\mypara{Explicit Representations.}
In traditional computer graphics pipelines, 3D shapes are predominantly represented as polygonal meshes (most commonly triangle meshes).
Meshes encode both the geometry through the spatial coordinates of vertices on the shape surface, as well as the topology connecting those vertices.
Material properties of the shape's surface such as albedo color, specularity, reflectance etc. can be represented through values at the vertices or more densely through a texture map that is mapped onto the surface, typically through the use of barycentric coordinate-based interpolation on the triangles constituting the mesh.
Given a 3D surface mesh representation, generation of a point cloud is possible through sampling of the mesh surface, where sampled point coordinates are in $\mathbb{R}^3$, and additional information such as the surface normal vector, surface color value etc. can also be extracted.
Through the process of voxelization, it is also possible to generate voxel grid representations of occupancy, color, or signed distance from the nearest surface.
Point clouds and dense voxel grid representations are common choices for encoding 3D shape structure in neural architectures.
In contrast, triangle meshes are the dominant representation for exchange and storage in 3D content databases, rendering, and other computer graphics pipeline operations.

\mypara{NeRF \citep{mildenhall2020nerf}.} The Neural Radiance Field (NeRF) represents a 3D scene as a continuous function $\sigma, c = \text{NeRF}(xyz,\mathbf{d})$, where $xyz$ is a query point within the 3D scene, $\mathbf{d}$ consists of the camera viewing direction and $\sigma, c$ are the density and color. To render a pixel from a ray $r$ the following equation is used:
\begin{equation}
    \hat{C}(r)=\sum_{i=1}^{K}Tr_i\alpha_ic_i\text{, where } \alpha_i=1-\exp(-\sigma_i\delta_i)\text{, and } Tr_i=\prod_{j=1}^{i-1}(1-\alpha_j)
\end{equation}
Here $Tr_i$ is the transmittance, $\alpha_i$ is the alpha compositing value, $\delta_i$ is the length between neighboring samples on the ray and $\hat{C}(r)$ is the rendered color for ray $r$. Here we mention some notable works that are used by methods in this survey. Voxel grid NeRF models~\citep{sun2022direct, chen2022tensorf, fridovich2022plenoxels, karnewar2022relu} store learnable explicit/implicit features in a voxel grid structure for fast learning due to the smaller or outright removal of the MLP predicting density and color. InstantNGP~\citep{mueller2022instant} uses multi-resolution hash tables for storing learnable features. Mip-NeRF~\citep{barron2021mip} uses conical frustums as opposed to rays for more accurate rendering. Mip-NeRF 360~\citep{barron2022mipnerf360} further add changes to the scene parameterization for better reconstruction of unbounded scenes. \citet{xie2022neural} offers a survey of neural fields and their applications.

\mypara{DMTet \citep{shen2021dmtet}.} DMTet uses a hybrid representation scheme with an explicit tetrahedral grid~\citep{gao2020learning} with signed distance field (SDF) values at each vertex to implicitly represent the underlying isosurface. Adopting their notation, the tetrahedral grid $T$ containing the vertices $V_T$ form $(V_T, T)$. The tetrahedrons in the grid $T_k\in T$ consists of four vertices $\{v_{a_k}, v_{b_k}, v_{c_k}, v_{d_k}\}$, where there are $K$ tetrahedrons in total. Given the SDF values at each vertex $\{SDF(v_{a_k}), SDF(v_{b_k}), SDF(v_{c_k}), SDF(v_{d_k})\}$ we can calculate the zero crossing via interpolation on edges where sign changes occur. The surface within the tetrahedra can then be extracted, this process is referred to as Marching Tetrahedra. Deformation vectors are also encoded for each vertex, this allows for fitting 3D objects with even finer detail. As we will see for works using DMTet for text to 3D, the SDF and deformation values can either be optimized explicitly or predicted from a neural network. Along with color information for each vertex the extracted mesh is often rendered with a differentiable rasterizer~\citep{laine2020modular} for optimization.

\edited{
\mypara{3D Gaussians.}
\citet{kerbl20233d} introduced blended 3D Gaussians for representing scenes.
Each Gaussian is parameterized by a position, scaling vector and, rotation as a unit quaternion.
The position corresponds to the mean of the Gaussian, and the covariance matrix is formed from the scale and rotation.
Despite its simplicity, this representation has been shown to be effective at accurately and efficiently modeling complex scenes and has been adopted by more recent work~\citep{tang2023dreamgaussian,yi2023gaussiandreamer,chen2023text}.
}

\edited{
\mypara{Structured Representations.}
It is also possible to represent a 3D shape as a combination of primitive parts.
The parts can be grouped in a hierarchical manner, or generated via a program.
This constitutes a promising direction for future work on text-to-3D generation: leveraging structured representations as used by neurosymbolic methods (see \citet{ritchie2023neurosymbolic} for a recent survey).
}

\subsection{Deep Generative Models} 
\label{sec:bg-generative}

A goal of text-to-3D shape generation is to generate potentially a variety of different shapes given a single text query.
Thus, a fundamental component in text-to-3D shape generation is the use of a generative model.
A generative model models the joint probability distribution and can be sampled from to obtain diverse samples.
In text-to-3D generation, the output space can be a 3D representation, a latent vector that is then transformed into a 3D representation via a shape decoder, or a 2D image that is used to generate a 3D shape.
In addition to the above, it is also possible to use a text-to-image generative model to guide the optimization of a 3D representation.
There are a number of popular families of generative models with different properties including auto-regressive models, GANs, VAEs, normalizing flow models, and diffusion models.
For a more detailed discussion \citet{shi2022deep} provide an excellent overview of the types of generative models and how they can be used in generation of 3D output.
Here, we elaborate on diffusion models since they Are an important component of recent work on text-to-3D generation that does not require 3D data.

\mypara{Diffusion Models.}
\citet{sohl2015deep, ho2020denoising} are generative models that model the distribution of the training data $x_0\sim q(x_0)$ by integrating latent variables $x_{1:T}$ by $p_\phi(x_0)=\int p_\phi(x_{0:T})dx_{1:T}$. Here we follow the notation used by DDPM~\citep{ho2020denoising}. The distribution of the latent variables is defined through a forward and reverse process using Markov Chains. The forward process involves corrupting a data sample $x_0$ at each time step according to a noise schedule until $x_T$ is just Gaussian noise $x_T\sim\mathcal{N}(0,\mathbf{I})$. The formal definition is as follows:
\begin{equation}
    q(x_{1:T}|x_0)=\prod^{T}_{t=1}q(x_t|x_{t-1})
\end{equation}
\begin{equation}
    q(x_t|x_{t-1})=\mathcal{N}(x_t;\sqrt{1-\beta_t}x_{t-1},\beta_t\mathbf{I})
\end{equation}
Here $\beta_t$ is a predefined variance schedule. Note that we can obtain the latent variable $x_t$ from $x_0$ by sampling $q(x_t|x_0)=\mathcal{N}(x_t;\sqrt{\bar{\alpha_t}}x_0,(1-\bar{\alpha_t})\mathbf{I})$, where $\alpha_t=1-\beta_t$ and $\bar{\alpha_t}=\prod_{s=1}^{t}\alpha_s$. The reverse process can similarly be defined through Gaussian transitions:
\begin{equation}
    p_\phi(x_{0:T})=p(x_T)\prod^{T}_{t=1}p(x_{t-1}|x_t)
\end{equation}
\begin{equation}
    p_\phi(x_{t-1}|x_t)=\mathcal{N}(x_{t-1};\mathbf{\mu}_\phi(x_t,t),\Sigma_\phi(x_t,t))
\end{equation}
In DDPM, by setting $\Sigma_\phi(x_t,t)=\sigma_t^2\mathbf{I}$ and $\mathbf{\mu}_\phi(x_t,t)=\frac{1}{\sqrt{\alpha_t}}(x_t-\frac{\beta_t}{\sqrt{1-\bar{\alpha_t}}}\epsilon_\phi(x_t,t))$, where $\epsilon_\phi$ is a neural network used to predict the unit variance noise $\epsilon$ used to corrupt $x_0$. We can supervise the training with a simplified noise residual objective:
\begin{equation}
    \E_{t,x_0,\epsilon}[w_t\|\epsilon-\epsilon_\phi(\sqrt{\bar{\alpha_t}}x_0+\sqrt{1-\bar{\alpha_t}}\epsilon,t)\|^2]
\end{equation}
where $w_t$ is a weighting term. The neural network $\epsilon_\phi$ is usually implemented as a UNet~\citep{ronneberger2015u}. After training the network parameters $\phi$ with the above loss, we can generate samples by first sampling $x_T\sim\mathcal{N}(0,\mathbf{I})$ then using the reverse process to get $x_0$. Note another parameterization of this loss in \citet{karras2022elucidating}:
\begin{equation}
    \E_{\sigma_t,\mathbf{y},\mathbf{n}}[\lambda(\sigma_t)\|D(\mathbf{y}+\mathbf{n};\sigma_t)-\mathbf{y}\|^2_2]
\end{equation}
where the denoiser is defined as $D(x;\sigma_t)=x-\sigma\epsilon_\phi(x,t)$, $\mathbf{y}$ is a sampled point from the training data and $n\sim\mathcal{N}(0,\sigma_t^2\mathbf{I})$ is the noise. We encourage readers to read \citet{karras2022elucidating} for more details as it brings recent diffusion methods under the same framework and discuss different hyperparameters choices.

\subsection{Guidance Models}
\label{sec:bg-guidance}

\mypara{CLIP \citep{radford2021learning}.}
The CLIP model consists of a text encoder $\EncCLIPText$ and image encoder $\EncCLIPImage$ which project text and images onto an aligned latent space trained by minimizing the contrastive loss on 400 million image text pairs.
The text and image embeddings are obtained through $\EmbCLIPText=\EncCLIPText(\InputText)$ and $\EmbCLIPImage=\EncCLIPImage(\InputImage)$, where $\InputText$ and $\InputImage$ are the input text and image respectively and $\EmbCLIP\in R^{z}$ is the embedding in the aligned latent space.
Finally, we can calculate the similarity between a pair of image and text as $\nicefrac{\EmbCLIPText\cdot\EmbCLIPImage}{\|\EmbCLIPText\|\|\EmbCLIPText\|}$.

\mypara{T2I Diffusion Models.}
Here we briefly discuss text to image (T2I) diffusion models used by the following works in this survey to provide a 2D prior for generating 3D objects.
Earlier works~\citep{sohl2015deep, song2020score, dhariwal2021diffusion} uses an additional classifier which is gradient conditioned on the noisy image and label to guide the sampling process.
More recent T2I diffusion models use classifier-free guidance~\citep{ho2022classifier} which is formulated as:
\begin{equation}
    \hat{\epsilon}_{\phi}(x_t, c)=(1+\omega)\epsilon_\phi(x_t;t,c)-\omega\epsilon_\phi(x_t;t,\varnothing)
\end{equation}
Here $c$ is the conditioning variable, which is the input text in the case of T2I models.
The predicted noise at each step involves the conditional and unconditional scores with weighting $\omega$ and $\varnothing$ is a null prompt.
The text conditioning can be incorporated through cross attention layers in the UNet network~\citep{rombach2022high, saharia2022photorealistic, balaji2022ediffi}.
eDiff-I~\citep{balaji2022ediffi} and Imagen~\citep{saharia2022photorealistic} follow a cascaded approach where the diffusion model is trained at a lower resolution, with cascading super resolution models to upsample the produced images.
Stable Diffusion~\citep{rombach2022high} uses a latent diffusion model (LDM) architecture that first trains to compress images to a lower resolution latent space with an autoencoder.
Then, this approach fits a diffusion model in the latent space learned in the first stage.
Due to its open source nature, Stable Diffusion is a popular choice used by the majority of works.

\section{Text-to-3D using 3D Data}
\label{sec:method-3ddata}

\begin{figure}
\centering
\includegraphics[trim=0 0 0 0,clip,width=\linewidth]{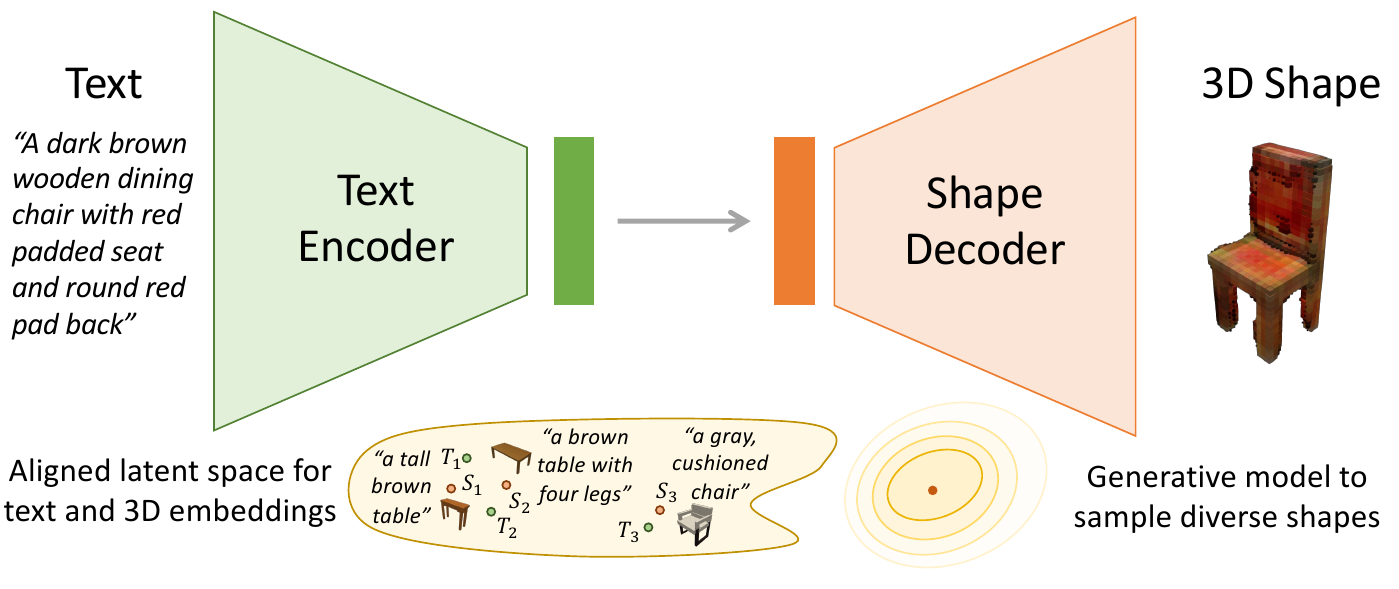}
\caption{
Illustration of key components in text to 3D generation models.
A \emph{text encoder} takes the text and produces an embedding in a latent space that is used to condition a generative model to sample a latent shape code.
The latent shape code is then passed through a \emph{shape decoder} to generate a 3D shape.
Typically, the text and 3D latent spaces are aligned so the text embedding can be directly used as the shape embedding.
The generative model is used to allow for sampling of diverse shapes.
Text-to-3D models are characterized by the choice of text encoder, shape decoder, shape representation, text-3D alignment method, and how the generative model is integrated to allow  sampling of diverse shapes.
}
\label{fig:overview-aligned}
\end{figure}

At a high-level, text-to-3D methods require the following components (see \Cref{fig:overview-aligned}):
\begin{itemize}
\item \emph{Text encoder.} Encodes text into an embedding space.
\item \emph{Shape decoder.} Generates a 3D shape from a latent vector. When paired text-3D data is available, this can also be trained to generate (or decode) from the encoded text embedding.
\item \emph{Joint text-shape embedding.} When paired text-3D data is available, this can be learned directly.   
\item \emph{3D generative model.} This generative model can be used to generate a latent vector for the shape decoder, images from which shapes are then generated, or directly to generate diverse shapes.
\end{itemize}

Note that it is possible to train the text encoder and shape decoder separately on text only and 3D only data.
The shape decoder is often trained as part of the shape autoencoder with just 3D data.
When paired text-3D data is available, the joint text-shape embedding specifying the alignment can be trained directly.
When there is no aligned text-3D data, image embeddings are typically used to bridge the two modalities.
Specifically, pretrained vision-language models (where the text-image embeddings are already aligned) are used, and the shape encoder is trained to align shape embeddings into the same space, with the shape decoder then trained to decode from that space.

We first discuss paired text-3D data methods (\paired) in \Cref{sec:method-3ddata-paired}, and then discuss unpaired text-3D data methods (\unpaired) in \Cref{sec:method-3ddata-unpaired}.

\subsection{Paired Text to 3D (\paired)}
\label{sec:method-3ddata-paired}

Traditional supervised approaches are based on the assumption that training data providing paired text and 3D samples is available.
Recent work based on this assumption can produce models that are of very high quality.
However, these approaches are typically unable to generate objects outside of the dataset and the quality of the 3D objects is highly dependent on the training dataset used.
A summary of paired text-to-shape datasets can be found in \Cref{tab:paired-data-summary}.

\begin{table}[t]
\resizebox{\linewidth}{!}
{
\begin{tabular}{@{} l c rc cc @{}}
\toprule
Dataset & Contrastive & Shapes & Categories & Text & Text source  \\
\midrule
Text2Shape~\cite{chen2019text2shape} & no & 15K  & 2 & 75K & crowdsourced \\
PartIt~\cite{hong2021vlgrammar} & yes & 10K & 4 & 10K & crowdsourced \\
\edited{Cap3D~\cite{luo2023scalable}} & \edited{no} & \edited{550K-785K} & \edited{many} & \edited{---} & \edited{generated} \\
OpenShape~\cite{liu2023openshape} & no & 876K & many & --- & generated \\
\notavail{Point-E~\cite{nichol2022point}} & no & >1M & many & --- & --- \\
\midrule
ShapeGlot~\cite{achlioptas2019shapeglot} & yes & 5K & 1 & 79K & crowdsourced\\
SNARE~\cite{thomason2022language} & yes & 8K & 262  & 50K & crowdsourced \\
ShapeTalk~\cite{achlioptas2023shapetalk} & yes & 36K & 30 & 536K & crowdsourced \\
\bottomrule
\end{tabular}
}
\caption{Datasets of paired text and shape.
Gray text indicates dataset is not open-sourced.
The `---' symbol indicates the statistics are unavailable.
\edited{Cap3D offers various 3D object-caption paired data versions of varying quality depending on filtering.}
}
\label{tab:paired-data-summary}
\end{table}

\Cref{tab:methods-3DPT} summarizes the methods in terms of text encoder, text-to-3D alignment approach, and generative model used.
We categorize supervised methods into three primary classifications.
The first encompasses conventional techniques~\citep{chen2019text2shape, liu2022towards} that learn an aligned space between text and 3D using modality-specific encoders.
These latents are subsequently inputted into a 3D object decoder to produce the final shape.
The subsequent two categories first employ an auto-encoder for 3D shapes.
A separate network is then utilized to learn a prior based on the latent space derived from the initial stage.
Text conditioning is integrated through various mechanisms within the prior network, producing latents conditioned on the input text.
The key distinction between these latter two categories lies in their modeling approach: one leverages autoregressive models for prior learning~\citep{mittal2022autosdf, fu2022shapecrafter}, while the other employs diffusion models~\citep{cheng2023sdfusion, li2023diffusion, zhao2023michelangelo, li20233dqd, jun2023shap}.  

Much of the work in this area is focused on improving 3D generative modeling, with the text as an illustrative conditioning input along side other potential conditioning input. 

\subsubsection{Paired Text-to-shape Datasets}

There are few datasets that provide both 3D shapes and natural language text descriptions.
\emph{Text2Shape}~\citep{chen2019text2shape} provided the first paired text and 3D dataset based on the text and chair models from ShapeNet~\citep{chang2015shapenet}.
Descriptions provide information about both color and shape.
\emph{ShapeGlot}~\citep{achlioptas2019shapeglot} and \emph{ShapeTalk}~\citep{achlioptas2023shapetalk} provided discriminative text that selected one object from multiple objects.
However, as noted by \citet{luo2022neural}, this style of text data omits important information that is shared between the three objects (e.g. the object category) and is not suitable for aligning text and 3D spaces for 3D shape generation.
This data is however, appropriate for shape editing.
Many of the shape datasets come from 3D model repositories that contain text information (such as the name of the asset, product catalogue descriptions, etc).
Such text information can be found in datasets such as ABO~\cite{collins2022abo}, Objaverse~\cite{deitke2023objaverse}, and Objaverse-XL~\cite{deitke2023objaversexl}.
However, this text information is often noisy, uninformative, and for better or worse, in multiple languages.
\emph{OpenShape}~\citep{liu2023openshape} \edited{and Cap3D~\citep{luo2023scalable}} illustrated the use of large-language models (LLMs) for filtering and generating more informative text descriptions for shapes \edited{on the large scale Objaverse~\cite{deitke2023objaverse} dataset}.
\Cref{tab:paired-data-summary} summarizes these paired datasets.

\begin{table*}
\resizebox{\linewidth}{!}
{
\begin{tabular}{@{} l crccccccr @{}}
\toprule
& & & & & \multicolumn{5}{c}{\edited{Model}} \\
\cmidrule(lr){6-10}
Method & \edited{Dataset} & 3D rep & Color & Images & Text &  \edited{Text-to-3D align} & \edited{Shape} & \edited{Gen model} & \edited{Gen space} \\
\midrule
Text2Shape\citep{chen2019text2shape} &  \edited{Text2Shape} & voxels ($32^3$)& yes & no & CNN + GRU & metric + label by assoc. & --- & WGAN & \edited{voxels} \\
TITG3SG\citep{liu2022towards} &  \edited{Text2Shape} & implicit occ. & yes & no & BERT & cyclic loss & \edited{AE} & IMLE & \edited{latent} \\
AutoSDF\citep{mittal2022autosdf} & \edited{ShapeGlot} & T-SDF ($64^3$) & no & no & BERT & --- & VQ-VAE & \edited{Autoregressive} & \edited{latent} \\
SDFusion\citep{cheng2023sdfusion} &  \edited{Text2Shape} & T-SDF ($64^3$/$128^3$) & texture & no & BERT & --- & VQ-VAE & \edited{Diffusion} & \edited{latent} \\
Diffusion-SDF\citep{li2023diffusion} &  \edited{Text2Shape} & T-SDF ($64^3$) & no & no & CLIP & --- & VAE & \edited{Diffusion} & \edited{latent} \\
3DQD\citep{li20233dqd} & \edited{ShapeGlot} & T-SDF & no & no & CLIP & --- & P-VQ-VAE & \edited{Diffusion} & \edited{latent} \\
Shap-E\citep{jun2023shap} & \edited{Internal Dataset} & STF & yes & multi & CLIP & --- & \edited{Transformer + NeRF} & \edited{Diffusion} & \edited{latent} \\
Michelangelo\citep{zhao2023michelangelo} &  \edited{ShapeNet + templates} & pt cloud / occ. & no & multi & CLIP & contrastive loss & SITA-VAE & \edited{Diffusion} & \edited{latent} \\
\edited{SALAD\citep{koo2023salad}}  & \edited{ShapeGlot} & \edited{3D Gaussian + impl. occ} & \edited{no} & \edited{no} & \edited{LSTM} & \edited{---} & \edited{AD} & \edited{Casc. diffusion} & \edited{exp. + latent} \\
\edited{ShapeScaffolder\citep{tian2023shapescaffolder}} & \edited{Text2Shape} & \edited{structured impl. occ} & \edited{yes} & \edited{no} & \edited{BERT + graph} & \edited{MSE + part-node attn} & \edited{AE} &  \multicolumn{2}{c}{\edited{None - hierarchical decoding}}  \\  
\bottomrule
\end{tabular}
}
\caption{
\edited{Methods that use paired 3D data with text (\paired).}
Early work~\cite{chen2019text2shape} did not rely on pretrained models and trained everything from scratch.
Later models initially leverage pretrained language models (e.g. BERT), and eventually pretrained vision-language models (e.g. CLIP/GLIDE) for encoding the text.
We use `---' to indicate that there was no \emph{extra} text-to-3D alignment, and that the alignment is done as part of the training of conditional generation with text input.
In the color column, `texture' indicates the work proposes a way to apply texture to the generated 3D shape.
The `Images' column indicates whether the method uses single or multi-view images for training or aligning image to shape models.
\edited{The `Shape' column indicates how encoding-decoding into/from the 3D shape latent space is trained, typically with a 3D autoencoder (AE) or variational autoencoder (VAE), with a few works~\cite{koo2023salad} using auto-decoding (AD) as introduced in DeepSDF~\cite{park2019deepsdf} (where only the decoder is trained).
Early work~\cite{chen2019text2shape} did not train to encode into the shape latent space separately.
In Shap-E~\cite{jun2023shap}, the latent code is used as parameters for a NeRF/STF MLP, and a transformer was trained to project point cloud and multi-view data into the latent space using rendering losses for NeRF and STF.
We indicate the type of Generative model (`Gen model') used and whether the model is operating in the latent space or not (`Gen space').
Note that WGAN and IMLE (used in \cite{chen2019text2shape, liu2022towards}) require only a single forward pass for sampling and decoding the 3D shape during inference.
Newer methods first require autoregressive or diffusion sampling in a latent space then decoding with a decoder model which usually means a longer inference time.
We can succinctly refer to autoregressive models and diffusion models in latent space as LAM and LDM respectively (see \Cref{fig:overview-lamldm}), but here we explicitly indicate the generative model and space for clarity.}}
\label{tab:methods-3DPT}
\end{table*}

\subsubsection{Notable Examples of \paired Methods}

\mypara{GAN-based.}
The pioneering work \emph{Text2Shape}~\citep{chen2019text2shape} was the first work to use deep generative models to generate shapes from text.
They modeled the shapes as dense voxel grids representing occupancy and RGB color.
Their text to 3D method involves two training stages.
The first stage uses learning by association~\cite{haeusser2017learning} combined with metric learning~\cite{sohn2016improved} to learn a shared representation between shapes and text descriptions.
Then, a voxel-based conditional Wassertein GAN~\citep{mirza2014conditional, arjovsky2017wasserstein, gulrajani2017improved} model is used to learn to generate shapes from input text prompts.
When training the GAN, the critic judges both whether the generated shape is `realistic' and whether it matches the text.

\mypara{Auto-encoder with IMLE.}
\citet{liu2022towards} proposed a three stage training process.
The first stage learns an auto-encoder model to reconstruct the object occupancy and color grid, where the latent space is decomposed into shape and color features.
In the second stage, a text encoder is learned to output features in the same decomposed latent space of the first stage from the paired text data.
In addition, another decoder leveraging word and local point feature attention layers are learned for different local features based on point coordinates.
Finally, to enable diverse generations from text prompts a generator outputs different shape and text features given input noise vectors and a source shape and text feature.
IMLE~\citep{li2018implicit} is used to guarantee that each ground truth shape has corresponding samples from the generator.

\mypara{Autoregressive Prior.}
\emph{AutoSDF}~\citep{mittal2022autosdf} introduces the use of a Patch-wise encoding VQ-VAE~\citep{van2017neural}, termed P-VQ-VAE to encode 3D shapes characterized by truncated signed distance function (T-SDF) by locally encoding patches independently into discrete encodings, arranged in a grid structure, followed by a joint decoding process.
They first train the P-VQ-VAE on 3D shapes from ShapeNet.
Then a non-sequential autoregressive model is proposed to learn a prior over the discrete latent space learned from the first stage.
For language conditioning, an auxiliary network is trained to predict latents based on a given text prompt, leveraging text-shape pairs from ShapeGlot~\citep{achlioptas2019shapeglot} for training.
Beyond this, their model is also capable of performing other tasks like shape completion and 3D reconstruction.

\textit{ShapeCrafter}~\citep{fu2022shapecrafter} enhances the work of AutoSDF by introducing the Text2Shape++ dataset, an advancement of the original Text2Shape.
This dataset breaks down text prompts into phrase sequences and calculates the similarity between these sub-sequences and shapes to establish a many-to-many shape correspondence.
This approach supports recursive 3D shape modeling during generation.
Although ShapeCrafter employs the same P-VQ-VAE from AutoSDF for latent shape representation learning, it uses a recursive shape generation training process.
Here, the phrase sequence is recursively fed to the autoregressive model at each time step, leading to progressive shape generation.

\begin{figure}
\centering
\includegraphics[width=\linewidth]{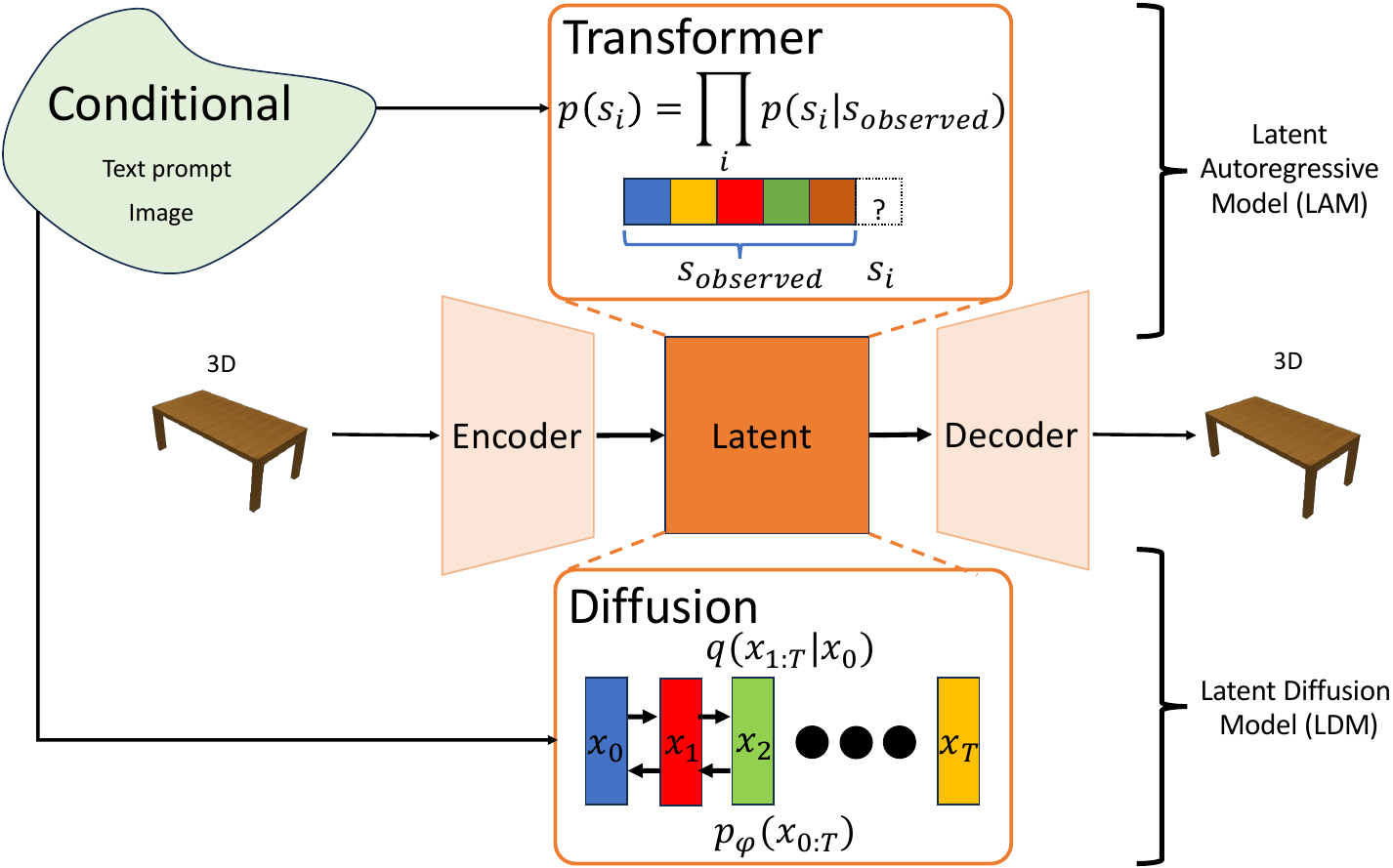}
\caption{Illustration of the difference between latent autoregressive models (LAM) and latent diffusion models (LDM). LAM utilizes an autoregressive prior over a learned latent space whereas LDM uses a diffusion model.
LDMs have demonstrated improved performance in text-to-3D shape generation.
}
\label{fig:overview-lamldm}
\end{figure}

\mypara{Diffusion Prior.}
Recent advances in diffusion models have shifted attention from autoregressive models to diffusion-based techniques for modeling the latent space.
This trend was notably driven by Latent Diffusion Models (LDMs)~\citep{rombach2022high}.
\citet{li20233dqd} pinpointed several limitations of autoregressive models, including:
1) error accumulation during sampling;
2) suboptimal generation order due to the processing direction inherent in transformers; and
3) a tendency for transformers to mode-collapse without adequate noise injection, resulting in reduced diversity.
Progress in text-to-image diffusion models has streamlined the integration of conditional text into the diffusion process.
For instance, the use of domain-specific encoders and a cross-attention mechanism in \citet{rombach2022high} can be adeptly employed for text-guided 3D generation.
Similarly, techniques like Img2Img~\citep{meng2021sdedit} have been used for text-driven editing and manipulation.
\Cref{fig:overview-lamldm} illustrates the contrast between latent diffusion models (LDM) and latent autoregressive models (LAM).

\textit{SDFusion}~\citep{cheng2023sdfusion} similarly employs a two-stage approach for 3D shape generation.
Initially, it trains a 3D VQ-VAE model on T-SDF representations using ShapeNet.
Subsequently, a 3D UNet is utilized to learn the LDM for the latent space established in the first stage.
Conditional generation with different modalities is incorporated following \cite{rombach2022high} with modality-specific encoders and a cross-attention mechanism within the 3D UNet.
During inference, classifier-free guidance~\citep{ho2022classifier} is used for conditional shape sampling.
Additionally, the model integrates Score Distillation Sampling (SDS)~\citep{poole2022dreamfusion} to facilitate 3D shape texturing.
Evaluation on the Text2Shape and ShapeGlot datasets reveals that SDFusion outperforms autoregressive methods such as AutoSDF.

\emph{Diffusion-SDF}~\citep{li2023diffusion} employs a patch-based VAE~\citep{kingma2013auto} model to encode local patches derived from the SDF of shapes, utilizing a patch joint decoder for reconstruction.
For its LDM, a 3D UNet-based architecture named UinU-Net is introduced.
This architecture incorporates an inner network using $1\times1\times1$ convolutions to focus on individual patch information.
The inner network is integrated with the primary UNet via skip connections.
Text-conditioned generation parallels the approach of SDFusion.
For text-driven shape completion, the model leverages inpainting mask-diffusion methods~\citep{lugmayr2022repaint, rombach2022high}.
Furthermore, diffusion based image manipulation techniques~\citep{chandramouli2022ldedit, kim2022diffusionclip} facilitate text-guided shape completion.

\emph{3DQD}~\citep{li20233dqd} follows the use of P-VQ-VAE in AutoSDF to learn encodings for 3D shapes.
Distinctly, 3DQD's LDM model learns a discrete diffusion process directly on the one-hot encodings of the VQ-VAE's codebook indexes with transition matrices during the diffusion process.
Given the noise and categorical corruption introduced by this discrete diffusion, a Multi-frequency Fusion Module (MFM) is integrated at the end of the denoising transformer to mitigate high-frequency anomalies in token features.

\emph{Shap-E}'s 3D encoder~\citep{jun2023shap} accepts both point cloud data and multi-view renderings of a 3D object.
The input data is integrated through cross-attention and subsequently processed by a transformer backbone.
This backbone produces a latent representation which acts as the weight matrices for an MLP.
Notably, this transformer functions as a hypernetwork~\citep{ha2017hypernetworks}.
The MLP is designed to predict STF outputs.
These outputs include three distinct branches: an SDF branch, a color branch, and a density branch.
The latter two are used for the NeRF representation.
The SDF values facilitate the generation of a mesh using the marching cubes algorithm, which can then be rendered with the color data.
As a result, Shap-e's decoder can produce both NeRF and mesh outputs.
For training purposes, photometric losses are used with the images rendered from both NeRF and mesh representations.
The training dataset comes from \citet{nichol2022point}, boasting several million 3D assets paired with textual descriptions.
The authors expand upon this dataset with roughly 1 million more 3D assets and an additional 120K captions.
Furthermore, their LDM model is trained to predict the MLP weight matrices, which are based on the latents obtained from the 3D encoders.
To incorporate text conditioning, the text embedding is prepended to the input of the transformer diffusion model.
Thanks to the large scale of the training dataset, Shap-E can generate a rich array of diverse 3D objects.
However, regrettably, this training dataset is not publicly available to the research community.

\emph{Michelangelo}~\citep{zhao2023michelangelo} introduces a methodology to learn an alignment between 3D shapes, text, and images prior to generation.
Initially, a transformer is used to encode the 3D point cloud, resulting in shape tokens.
Using frozen CLIP encoders~\citep{radford2021learning}, similar encoding processes are applied to both text and images, producing text and image tokens, respectively.
To ensure alignment across these three modalities, modality-specific projectors are used to process these tokens into distinct image, text, and shape embeddings.
The alignment of these embeddings is then done through a contrastive loss between the (shape, image) pairs and the (shape, text) pairs.
A reconstruction loss, based on the occupancy grid generated by the 3D decoder, is also employed.
The overarching network architecture is termed the Shape-Image-Text Aligned Variational Auto-Encoder (SITA-VAE).
After alignment, an Aligned Shape LDM (ASLDM) is trained to model the shape tokens.
The process of conditioning on text and image tokens follows the methodology presented by \citet{rombach2022high}.

\subsubsection{Structure aware text to shape generation}
\edited{
Most methods we described so far treat the output shape as an unstructured 3D representation.
An emerging line of work attempts to produce structured 3D shapes based on parts and part connectivity~\citep{luo2022neural,tian2023shapescaffolder}.  ShapeScaffolder~\cite{tian2023shapescaffolder} first uses shape data to pretrain a structured shape decoder that can take a global latent vector encoding (consisting of both shape and color latents) and hierarchically decode it into parts.
For text conditioning paired text-shape data is used to align a global text embedding to the shape and color latent spaces via MSE loss.
The text is also converted to a graph representing parts, their attributes, and relations.
Features for the parts and relations are extracted and attention between the text-based graph features and part features guides the hierarchical decoding process.  
Another recent work, SALAD~\cite{koo2023salad} proposed a two-phase diffusion approach.
In the first phase, the part structure is generated as a mixture of 3D Gaussians, and in the second phase a latent vector representing the shape details for each part is generated.
To condition the model on text, they concatenate the language feature and train the model using paired text-shape data from ShapeGlot~\cite{achlioptas2019shapeglot}.
There is also growing interest in generating shapes based on programs~\cite{jones2023shapecoder,yuan2023cadtalk}.
A recent work that connects text to programs is Neural Shape Compiler~\cite{luo2022neural}.
Translating text to programs that can generate 3D shapes is an interesting direction for future investigation. 
}

\begin{figure}
\includegraphics[width=\linewidth]{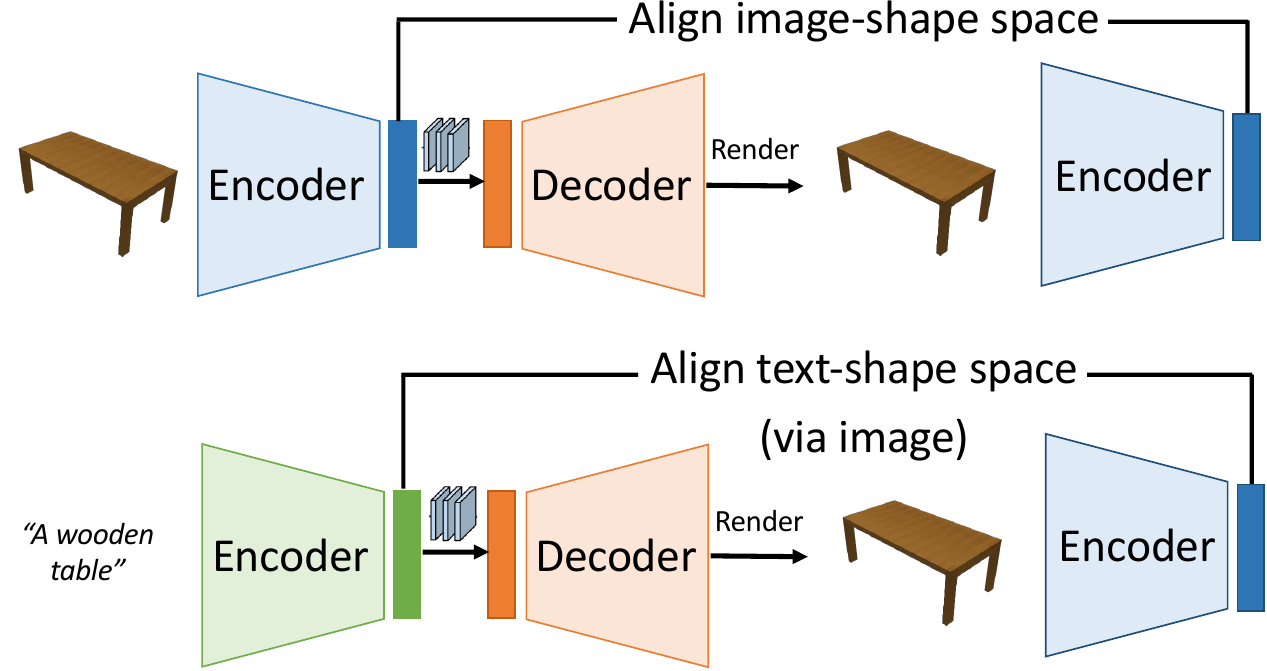}
\caption{
\edited{
In text-to-shape generation with unpaired data (\unpaired), an image-to-3D generative model is combined with pretrained text-image embeddings.  
Typically, 3D assets are first used to train a 3D autoencoder that takes a shape embedding to an 3D shape.
Large pretrained text-image embeddings are then used to align the text to 3D space, with a generative model that can sample from the 3D latent space conditioned on an input embedding.
The shape space is aligned to the pretrained text-image space by:
1) aligning the shape decoder to render from image embeddings e.g., as in CLIP-Forge and CLIP-Sculptor (top); or
2) ensuring 3D shapes generated from a text embedding have rendered embeddings that match the text embedding, e.g., as in TAP3D (bottom).
}
}
\label{fig:overview-unaligned}
\end{figure}

\subsection{Unpaired 3D Data (\unpaired)}
\label{sec:method-3ddata-unpaired}

\begin{table*}
\resizebox{\linewidth}{!}
{
\begin{tabular}{@{} l ccrcccc @{}}
\toprule
& & & & & &   \multicolumn{2}{c}{\edited{Model}} \\
\cmidrule(lr){6-8}
Method & \edited{Dataset} & 3D representation & Color & Images & Text & 
\edited{Text-to-3D alignment} & \edited{Generative model} \\
\midrule
CLIP-Forge\citep{sanghi2022clip} & \edited{ShapeNet} & voxels & no & multi & CLIP & CLIP + image / 3D align & Flow (latent) \\
CLIP-Sculptor\citep{sanghi2023clip} & \edited{ShapeNet} & voxels ($32^3$/$64^3$) & no & multi & CLIP & \edited{CLIP} & \edited{LAM} \\
ISS\citep{liu2022iss} & \edited{ShapeNet, CO3D} & DVR & yes & single & CLIP & CLIP + mappers + test time consistency loss & DVR \\
\rowcolor{verylightgray} TAP3D\citep{wei2023taps3d} & \edited{ShapeNet} & DMTet & yes & multi & CLIP & CLIP + mappers + align loss & GAN \\
\bottomrule
\end{tabular}
}
\caption{
\edited{Methods that use unpaired 3D data (\unpaired).}
The `Images' column indicates whether the method uses single or multi-view images for training or aligning image to shape models.
For `Text-to-3D alignment', all models use CLIP to align 3D objects during training with text prompts during inference.
Additional components may be further leveraged to better align image, 3D or text to bridge domain gaps in the CLIP latent space.
Note that TAP3D\cite{wei2023taps3d} (in light gray) trains an aligned text-shape space using generated templated text.}
\label{tab:methods-3DUT}
\end{table*}

In this section, we discuss methods where a database of 3D shapes is available for training a 3D generative model, but there is no paired text-to-3D data.
These methods are summarized in \Cref{tab:methods-3DUT}.
In this case, pretrained text-to-image models are used as a bridge to align the text and 3D embeddings.
Without paired 3D data, but access to a 3D dataset, it is possible to render 2D images from the 3D data and use the generated images as training data to train a generative model that goes from 2D image to 3D shapes.
Work in this space can either train their own image-to-3D model or leverage existing pretrained models. 
The generative model will typically make use of latent image embeddings that comes from a pretrained vision language model (e.g. CLIP).
Since CLIP has aligned text-and-image embeddings, this family of work assume using CLIP text-embedding directly will be sufficient.
However, recent work has identified that the pretrained embeddings have shortcomings and proposed ways to compensate.
\Cref{fig:overview-unaligned} provides an illustrative summary of this family of approaches.

\subsubsection{Notable Examples of \unpaired Methods}

\mypara{CLIP-Forge~\citep{sanghi2022clip}.}
Initially, a shape autoencoder is trained on the ShapeNet dataset to reconstruct 3D voxels.
This involves the encoder transforming voxels into embeddings, represented as $\EmbVoxel = \EncVoxel(\InputVoxel)$. Subsequently, both the embedding and voxel positions are inputted into a decoder occupancy network~\citep{mescheder2019occupancy}, denoted as $\DecVoxel$, to recreate the original shape.

The following stage uses a flow model~\citep{dinh2016density}.
This model is trained to process the embeddings, $\EmbVoxel$, in conjunction with CLIP image embeddings derived from multi-view renderings of the 3D objects.
These CLIP embeddings are represented as $\EmbCLIPImage=\EncCLIPImage(\InputImage)$, where $\InputImage$ are the rendered images.
The flow model maps inputs onto a Gaussian distribution.

During inference, multi-view images are substituted with text prompts.
These prompts are then encoded using CLIP to obtain the associated embeddings, represented as $\EmbCLIPText=\EncCLIPText(\InputText)$.
Given that the CLIP encoders are trained to map to a unified latent space, both the text embedding and a sample from the Gaussian distribution are inputted into the flow model in reverse to obtain the shape embedding, $\EmbVoxel$. Finally, the occupancy network decodes the final shape.

\mypara{CLIP-Sculptor~\citep{sanghi2023clip}.}
The concept of leveraging CLIP encoders that map to a shared latent space is also seen in CLIP-Sculptor.
However, CLIP-Sculptor employs a more powerful autoregressive model in lieu of a flow model.
During the initial training phase, two VQ-VAEs are trained at varying resolutions for reconstruction.
For the 3D shape, the voxel encoder transforms it into $\EmbVoxel_{r}=VQ(\EncVoxel_{r}(\InputVoxel_r))$, where $\EmbVoxel_{r}$ is the vector-quantized grid encoding for the input voxel and $r$ indicates the voxel resolution.
The lower resolution voxel grid is configured at $32^3$, while the higher resolution operates at $64^3$.
The shape is then reconstructed with the decoder $\hat{\InputVoxel_r}=\DecVoxel_r(\EmbVoxel_{r})$.

Similarly to CLIP-Forge, multi-view renderings of the 3D objects are encoded using CLIP, expressed as $\EmbCLIPImage=\EncCLIPImage(\InputImage)$.
Noise is introduced to these embeddings, resulting in the conditional vector $c$ for variation.
This vector is then processed by an MLP, which subsequently predicts the affine parameters of the layer normalization layers present in a coarse transformer with dropout.
This transformer is designed to autoregressively decode masked encodings under the constraint of conditionals, as denoted by $P(\EmbVoxel_{32}|\text{Mask}(\EmbVoxel_{32}), c)$.
In the subsequent training stage, a super-resolution (fine) transformer is trained to unmask high resolution encodings $\text{Mask}(\EmbVoxel_{64})$ conditioned on the predicted lower resolution encodings $\EmbVoxel_{32}$ from the coarse transformer through cross attention.

During the inference phase, the image embeddings are supplanted by text embeddings derived from user prompts.
Following this, the masked encoding undergoes an iterative decoding process, first by the coarse transformer and subsequently by the fine transformer, to produce the final encoding representing the desired shape.
Drawing inspiration from classifier-free guidance prevalent in diffusion techniques, CLIP-Sculptor frames the sampling with the equation: $\hat{P}_t(c)=P_t(0)+a(t)(P_t(c)-P_t(0))$.
Here, $P_t(c)$ is represented by $P(\EmbVoxel_{32,t}|\text{Mask}(\EmbVoxel_{32, t-1}), c)$ and $P_t(0)$ is given by $P(\EmbVoxel_{32,t}|\text{Mask}(\EmbVoxel_{32, t-1}), 0)$.
The function $a(t)$ starts with a high guidance scale, ensuring the shape adheres to the text-based constraints.
However, as the process advances, its value decreases to introduce greater diversity in the output.

\mypara{ISS~\citep{liu2022iss}.}
The two aforementioned papers operate with the assumption that the CLIP text and image spaces align seamlessly. However, the ISS authors find otherwise.
Specifically, they discovered significant distances between the embeddings of paired text and images.
When directly swapping the CLIP image encoder for the text encoder during inference, this discrepancy can lead to inconsistencies between the generated shape and its corresponding text.
Additionally, learning to generate shapes straight from the CLIP feature space causes a loss of details in the final 3D shape.
To counteract these issues, ISS introduces a two-phase method for aligning the feature spaces, enabling 3D shape generation guided by text without assuming perfectly matched text-shape pairs.

The initial alignment phase uses a pretrained single-view reconstruction model, specifically DVR~\citep{niemeyer2020differentiable}.
This comprises an encoder converting images to latent embeddings, and a decoder responsible for shape reconstruction.
They align the CLIP image features with the DVR network's latent features using a mapping network $M$.
This is achieved by minimizing the function $L_2(M(\EncCLIPImage(\InputImage)), \Enc^T_{DVR}(\InputImage))$.
Here, $\Enc^T_{DVR}$ represents the DVR decoder, $\InputImage$ are images rendered from the 3D object, and the CLIP features are normalized.
The decoder loss and other regularizing losses are optimized to refine the decoder network.

At the inference stage, to narrow the disparity between CLIP's image and text encoders, the mapping network $M$ undergoes additional fine-tuning. This is to minimize the CLIP consistency loss represented by $\langle\EncCLIPImage(\InputImage)\cdot\EncCLIPText(\InputText)\rangle$. Here, $\InputImage$ denotes images rendered from the 3D shape using the DVR decoder, while $\InputText$ is the provided text prompt. All CLIP features are also normalized. Acknowledging the constraints of the DVR model, they allow for further fine-tuning of the DVR decoder with the CLIP consistency loss for text-driven stylization.

\mypara{TAP3D~\cite{wei2023taps3d}} proposes a simple approach of adding text controls to an existing unconditional 3D generation model.
They utilize CLIP similarity as a loss to fine-tune parts of the original network to accept CLIP text embeddings as input.
Specifically, they use a GET3D~\cite{gao2022get3d} model pretrained on ShapeNet, a DMTet-based GAN.
The GET3D model takes as input two noise vectors and transforms them into intermediate latent codes using mapping networks for modeling the geometry and texture.
The mapping networks are the only parts of the model being fine-tuned.
First, to obtain text prompts for 3D shapes, they generate pseudo-captions by calculating nouns and adjectives that have high similarity with rendered images of the 3D objects.
They then construct the captions with template sentences.
The mappers are trained by maximizing the CLIP similarity between the pseudo-captions and rendered images of the generated object from GET3D.
An additional regularization loss aims to maximize the similarity between rendered images of generated objects and ground-truth object renderings.

\begin{figure*}
\includegraphics[width=\textwidth]{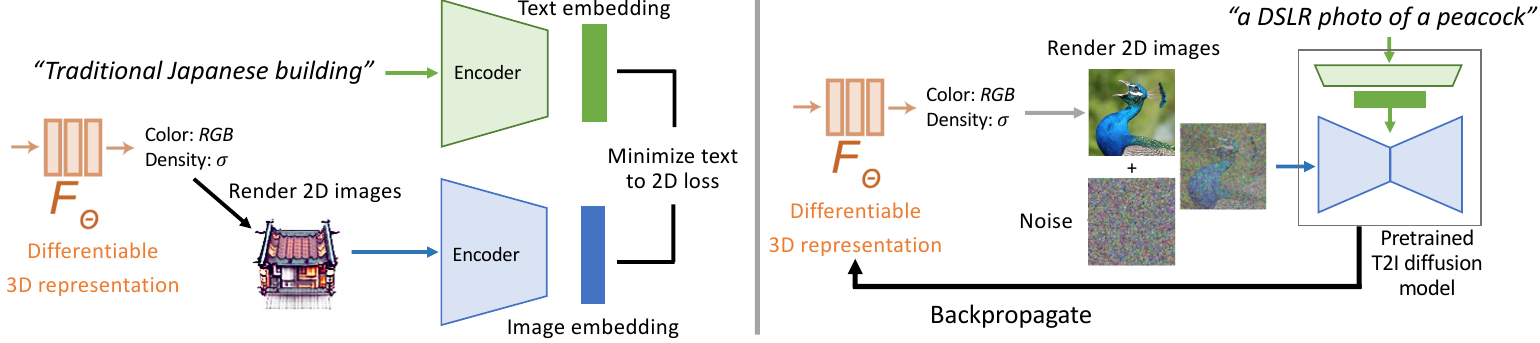}
\caption{
Illustration of the text-to-3D shape generation strategies that do not rely on 3D data.
Left: methods minimizing text to 2D loss using a pretrained vision-text embedding space such as CLIP.
Right: methods using pretrained image-to-text diffusion-based models to guide the training of a 3D representation.
These are per-prompt optimization strategies, leading to significant compute and time required per shape.
}
\label{fig:unsupervised-overview}
\end{figure*}

\section{Text-to-3D without 3D data (\notd)}
\label{sec:method-unsupervised}

Here we discuss the family of approaches designed for scenarios when either no 3D data is available, or reliance on such 3D data is avoided.
In this scenario, the typical strategy uses per-prompt optimization together with a differentiable renderer to optimize an underlying 3D representation such that 2D images that are compatible with the text prompt can be generated.
In this unsupervised setting, there are currently two popular sets of approaches:
1) maximizing the similarity of the prompt and rendered images using a pretrained vision-language joint embedding (\Cref{sec:method-unsupervised-clip}); or
2) using a pretrained text-to-image diffusion model to guide updates to the parameters of the 3D representation (\Cref{sec:method-unsupervised-diffusion}).
\Cref{fig:unsupervised-overview} illustrates these two sets of approaches.

A key design decision applying to both sets of approaches is the choice of 3D representation.
There are several popular choices that permit differentiable rendering:
1) NeRF-based model;
2) deforming triangular meshes; and
3) deep marching tetrahedra (DMTet).

\begin{table}
\resizebox{\linewidth}{!}
{
\begin{tabular}{@{} r rrrr @{}}
\toprule
Method & 3D Rep & Augmentation & Additional loss \\
\midrule 
DreamFields~\citep{jain2022zero} & MipNERF~\citep{barron2021mip} & BG (GN, CH, FT) & Transmittance  \\
CLIP-Mesh~\citep{mohammad2022clip} & mesh + normal + texture &  BG & Diff. prior + Lap reg\\
PureCLIPNeRF~\citep{lee2022understanding} & explicit/implicit NeRF & BG + diff + persp & Tr + TV + KL + Bg \\
\bottomrule
\end{tabular}
}
\caption{
Summary of unsupervised CLIP guidance methods.
These methods optimize the 3D representation so that the CLIP similarity between rendered images and text is maximized.
They also make use of additional augmentation and loss terms to help enforce the generation of more geometrically plausible objects.
For instance, DreamFields~\cite{jain2022zero} includes various background (BG) augmentations such as Gaussian noise (GN), checkerboard pattern (CH), and random Fourier textures (FT).}
\label{tab:methods-clip-guidance}
\end{table}

\subsection{Unsupervised CLIP Guidance}
\label{sec:method-unsupervised-clip}

In this class of methods, the parameters of a differentiable 3D representation are updated so that rendered images have high similarity to text prompt embeddings, as evaluated by a pretrained vision-language joint embedding (CLIP). 
This approach is challenging as training by just optimizing the CLIP similarity can be tricky and does not provide any signal for geometric consistency.
To counter this, work in this area proposes a variety of regularization and augmentation techniques.
\Cref{tab:methods-clip-guidance} provides a summary of methods and categorization along the axes of 3D representation, augmentation strategies, and additional regularization.

\subsubsection{Notable Examples}

\mypara{Dream Fields~\citep{jain2022zero}.}
Dream Fields is a pioneering work in this direction, first demonstrating the possibility of creating 3D shapes without relying on any 3D data.
While previous research utilized CLIP to bridge the language-shape gap in scenarios with available 3D data but lacking pairing with text, Dream Fields showcased the capability to guide the training of a 3D representation directly using pre-trained text-image CLIP embeddings.

To achieve this Dream Fields uses a NeRF as the backbone 3D representation.
Given a camera pose described by azimuth and elevation $(\theta, \phi)$, an image $\InputImage=NeRF(\theta, \phi)$ is generated using volumetric rendering.
The specific NeRF used is Mip-NeRF~\citep{barron2021mip}.
Then an input text prompt $\InputText$ guides the model training using the CLIP similarity loss, expressed as $\mathcal{L}_{CLIP}=-\EncCLIPImage(\InputImage)^T\EncCLIPText(\InputText)$.
A key challenge when using CLIP in this fashion is that the embeddings minimize a contrastive loss between images and text, and thus may prioritize texture learning over structural or spatial information.
Dream Fields shows that solely relying on this loss can lead to the generation of implausible 3D objects.
To counteract this, they propose multiple regularization techniques.
Initially, they incorporate a transmittance loss to eliminate floating density anomalies.
They also experiment with various background augmentations (e.g., Gaussian noise, checkerboard patterns, random Fourier textures).
These are alpha-composited onto the NeRF-rendered images, enhancing the coherence of the resulting 3D models.

\mypara{CLIP-Mesh~\citep{mohammad2022clip}.}
CLIP-Mesh similarly employs CLIP guidance.
However, they use a mesh representation. %
Notably, their training begins with a primitive geometric shape, such as a sphere.
This differentiates the approach from Text2Mesh~\cite{michel2022text2mesh}, which requires an initial object aligned with the semantic class of the input text prompt.

The primitive geometry's vertices dictate a subdivision surface, with both the normal map and texture maps initialized randomly.
The object is rendered from various viewpoints using a differentiable renderer, and these images are then processed through the CLIP image encoder.
The similarity loss with CLIP is subsequently calculated against the text embeddings.
Additionally, the CLIP-Mesh approach integrates a diffusion prior akin to DALL-E 2~\citep{ramesh2022hierarchical}.
Here, the text embeddings are mapped to image embeddings.
This is also leveraged to calculate similarity between the rendered images.
Their findings indicate that jointly employing these losses enhances generation quality.
Training is regularized using a Laplacian regularizer and random background augmentations (mirroring the Dream Fields approach).

\mypara{PureCLIPNeRF~\citep{lee2022understanding}.}
In the works discussed thus far, optimizing images under CLIP guidance often leads to incoherent results.
In PureCLIPNeRF~\citep{lee2022understanding}, various elements are examined to mitigate the adversarial generations caused by CLIP guidance.
These elements include different image augmentations, pretrained CLIP backbones, and both explicit and implicit NeRF architectures.
This work finds that the combination and selection of these factors influences the quality of the final output 3D shape results significantly.
By integrating augmentation strategies, CLIP backbones, and 3D representation design choices effectively, one can produce more consistent 3D objects.

\mypara{Dream3D~\citep{xu2023dream3d}.}
While the aforementioned regularizations aim to counteract adversarial generations, CLIP guidance remains susceptible to emphasizing textures or patterns from the input text prompt.
This often leads to inconsistencies in the geometry of the generated 3D object.
To address this, Dream3D~\cite{xu2023dream3d} introduces a 3D prior for initializing the CLIP guidance optimization.
This is achieved by fine-tuning a Stable Diffusion~\citep{rombach2022high} model using text and renderings from ShapeNet.
Another diffusion model is trained to translate the produced images into a 3D shape latent code.
This latent input is processed by an SDF-StyleGAN~\citep{zheng2022sdf} network to generate the 3D shape.
Ultimately, CLIP guidance refines this 3D model, ensuring it aligns more closely with the provided text.

\begin{figure}
\centering
\includegraphics[width=\linewidth]{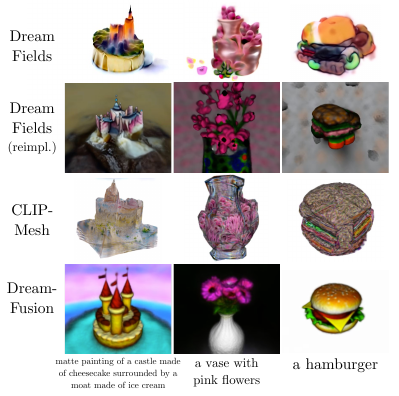}
\caption{Qualitative results from DreamFusion~\cite{poole2022dreamfusion} comparing against methods utilizing CLIP guidance (Dream Fields, CLIP-Mesh). DreamFusion produces higher quality 3D shapes. Visuals reproduced from \citet{poole2022dreamfusion}.
}
\label{fig:CLIPT2ICompare}
\end{figure}

\subsubsection{Discussion}

While methods in this section are not bounded to generating objects within the limits of a dataset like prior sections, the geometric quality of the output objects is still lacking in many cases.
This limitation is primarily because CLIP is trained with a contrastive objective, which can cause the image encoder to prioritize texture pattern extraction.
As a result, using CLIP for guidance may not effectively generate fine geometric detail on objects and can yield overly simplified geometry.
To address this issue, prior work has explored various augmentations that help regularize the training process, different CLIP architectures, as well as different 3D shape priors.
While these improvements can improve generated outcomes, the subsequent section will demonstrate that the alternative strategy using text-to-image diffusion models as guidance helps to increase the quality of the output 3D shape geometry as can be seen in \Cref{fig:CLIPT2ICompare}.
This strategy leads to objects that not only align more closely with more complex text prompts but also showcase detailed geometric and texture details.

\begin{table*}[t]
\resizebox{\linewidth}{!}
{
\begin{tabular}{@{} l ccrc @{}}
\toprule
Method & 3D representation & Guidance model & Loss \\
\midrule
DreamFusion~\citep{poole2022dreamfusion} & Mip-NERF 360~\citep{barron2022mipnerf360} & Imagen~\citep{saharia2022photorealistic} & SDS \\
SJC~\citep{wang2023score} & Explicit Voxel Grid NeRF & StableDiff~\citep{rombach2022high} & SJC \\
Prolific Dreamer~\citep{wang2023prolificdreamer} & 1) Instant-NGP~\citep{mueller2022instant} + 2) Dis. DMTet~\citep{shen2021dmtet, chen2023fantasia3d} & StableDiff & VSD \\
Magic3D~\citep{lin2023magic3d} & 1) Instant-NGP + 2) DMTet & 1) eDiff-I~\citep{balaji2022ediffi} + 2) StableDiff & SDS \\
TextMesh~\citep{tsalicoglou2023textmesh} & 1) SDF NeRF + 2) Mesh & 1) Imagen (SDS) + 2) StableDiff (texturing) & SDS \edited{+ Tex Ref} \\
Fantasia3D~\citep{chen2023fantasia3d} & Dis. DMTet & StableDiff & SDS \\
\edited{DreamGaussian~\citep{tang2023dreamgaussian}} & \edited{3D Gaussians} & \edited{StableDiff} & \edited{SDS + Tex Ref} \\
\edited{GSGEN~\citep{chen2023text}} & \edited{3D Gaussians} & \edited{StableDiff, Point-E} & \edited{SDS} \\
\edited{GaussianDreamer~\citep{yi2023gaussiandreamer}} & \edited{3D Gaussians} & \edited{StableDiff} & \edited{SDS} \\
\bottomrule
\end{tabular}
}
\caption{Summary of methods using unsupervised diffusion guidance, a sub-family of methods that do not require 3D data (\notd).  We organize these methods along the design choices of 3D representation, guidance model, and training loss. Some of the methods use a two stage approach to obtain a higher resolution mesh output or to separate geometry and texture (indicated by numerals 1 and 2 in each cell). \edited{Tex Ref indicates the use of additional texture refinement losses.}}
\label{tab:methods-diffusion-guidance}
\end{table*}

\subsection{Unsupervised Diffusion Guidance}
\label{sec:method-unsupervised-diffusion}

Much like the approach with CLIP guidance, we can bridge the gap between 3D shape being generated and the 2D guidance model by rendering images of the 3D shape from various camera perspectives.
Instead of optimizing the similarity between images and text, we utilize scores from the diffusion models as the guiding gradients.
The score, as determined by the denoising diffusion model $s_\phi(x_t, t)$, serves as an approximation to the score function $\nabla_{x_t}\text{log}p_{\alpha_t}(x_t|x)$~\citep{song2020score}.
Here, the score function $\nabla_{x_t}\text{log}p_{\alpha_t}(x_t|x)$ points towards areas of higher probability density.
In more detail, our objective is to ensure that images, irrespective of the angle from which the 3D model is rendered, gravitate towards high-probability zones outlined by the diffusion models.
Note that the noise function $\epsilon_\phi(x_t,t)$ is proportional to the score~\citep{robbins1992empirical}:
\begin{equation}
    -\frac{\epsilon_\phi(x_t,t)}{\sigma_t}=\frac{D(x_t;\sigma_t)-x_t}{\sigma_t^2}=s_\phi(x_t, t)
\end{equation}

\Cref{tab:methods-diffusion-guidance} summarizes the methods in this family, and categorizes them in terms of the underlying 3D representation, specific guidance model, and loss terms used during training.
In \Cref{sec:5-loss-formulation}, we highlight seminal papers that focus on formulating a loss around this concept.
\Cref{sec:5-representation-improvements} delves into works that leverage improved 3D representations for higher quality outputs.
A notable challenge with this formulation is the ``Janus Problem'', which is a consistency issue with the generated object.
This arises due to optimizing each rendered image without taking into account the entire object as well as inherent 2D biases in text-to-image (T2I) diffusion models.
To address this, \Cref{sec:5-janus-problem} presents studies that offer mitigation strategies and fine-tune T2I models on multi-view data, thereby injecting 3D information to counteract the 2D bias.
Another limitation of this framework is that we have to conduct training for each individual text prompt which can take up to several hours for high quality models.
However, the research in \Cref{sec:5-generative-model} suggests integrating this loss into a multi-object generative model, enhancing training efficiency.
This approach enables generation of multiple objects and interpolation between prompts during inference.

\subsubsection{Loss Formulation}
\label{sec:5-loss-formulation}

In this section, we highlight notable papers that introduce loss functions designed to harness 2D priors from text-to-image (T2I) diffusion models for generating 3D models from text.

\mypara{DreamFusion~\citep{poole2022dreamfusion}.}
The DreamFusion authors introduced the pioneering Score Distillation Sampling (SDS) loss.
This was used to optimize a NeRF model, specifically Mip-NeRF 360~\citep{barron2022mipnerf360}, parameterized by $\theta$.
It effectively integrates prior knowledge distilled from a pretrained T2I diffusion model Imagen~\citep{saharia2022photorealistic}, which is parameterized by $\phi$.
The optimization is driven by the following loss formulation:
\begin{equation}
    \nabla\mathcal{L}_{SDS}=\E[w(t)(\epsilon_\phi(x_t;y,t)-\epsilon)\frac{\partial x}{\partial \theta}]
\end{equation}
In this context, $x$ represents the image rendered from NeRF, while $y$ is the text prompt.
As previously mentioned, $\epsilon_\phi(z_t;y,t)$ relates to the score, signifying gradients that direct towards high-probability regions in the ambient space, conditioned on $y$.
The subtraction of the added noise $\epsilon$, serves as variance reduction.
This is crucial since the denoiser is conditioned on $x_t$, which exhibits noise levels distinct from those in the NeRF-generated image.
Furthermore, DreamFusion has shown that:
\begin{equation}
    \nabla_\theta\mathcal{L}_{SDS}(\phi, x=g(\theta)) = \nabla_\theta\E_t[\frac{\sigma_t}{\alpha_t}w(t)KL(q(x_t|g(\theta); y, t)\|p_\phi(x_t;y,t))]\label{eq:SDSKL}
\end{equation}
Optimizing the SDS loss aligns with minimizing the KL divergence between the noise-injected images from the NeRF network $g(\theta)$, and the probability densities learned by diffusion models conditioned on text prompt $y$.
For a detailed derivation, see the original paper.
Using the powerful priors from T2I models, DreamFusion achieves markedly improved results compared to prior methods employing CLIP.
It is worth noting that the more sophisticated shading model emphasizes geometric details, while a basic albedo model might induce ambiguities, impacting the quality of the object geometry.
A notable limitation is the ``Janus problem'' where models display multiple front faces, likely due to datasets used to train T2I models predominantly featuring front views of objects.

\mypara{Score Jacobian Chaining (SJC)~\citep{wang2023score}.}
Concurrently, Score Jacobian Chaining (SJC) formulates its approach by applying the chain rule to the score of the diffusion model.
This sidesteps the UNet Jacobian term present in DreamFusion, which originates from the diffusion training loss.
It is noteworthy that this term is excluded from the SDS loss, as empirical results demonstrated enhanced performance without it.
The SJC loss is as follows:
\begin{equation}
\nabla_\theta\mathcal{L}_{SJC}=E_n[\frac{D(x_\pi+\sigma n,\sigma)-x_\pi}{\sigma^2}]\frac{\partial x_\pi}{\partial \theta}
\end{equation}
The equation involves $\pi$, the camera pose for NeRF rendering, and $n$ are distinct noise samples employed for the Monte Carlo estimate.
Both the SDS and SJC formulations display comparable performance.
Their model employs Stable Diffusion~\citep{rombach2022high} as guidance and an explicit voxel grid NeRF model~\citep{chen2022tensorf, fridovich2022plenoxels, sun2022direct}.
By directly rendering in the Stable Diffusion latent space $R^{64\times64\times4}$, they accelerate training, sidestepping gradient computations through the encoder.
This technique is also advocated in Latent-NeRF~\citep{metzer2023latent}.

\mypara{ProlificDreamer~\citep{wang2023prolificdreamer}.}
A limitation of the SDS loss formulation is the tendency for generated objects to exhibit excessive smoothness and limited variation.
Prolific Dreamer theorizes that this stems from the mode-seeking formulation in \Cref{eq:SDSKL}.
Essentially, DreamFusion aims to fit a single plausible 3D model to a probability distribution within the diffusion model.
However, the distribution encompasses numerous objects with different identities fitting description $y$.
To address this, Prolific Dreamer approaches the issue through the lens of variational inference:
\begin{equation}
    \underset{\mu}{\min}KL(q_0^\mu(x_0|y) \| p_0(x_0|y))
\end{equation}
Here the goal is to minimize the KL divergence between images generated by a distribution $\mu$ comprised of multiple NeRF models (particles) and the pretrained diffusion model.
This leads to their ultimate formulation, termed the Variational Score Distillation (VSD) loss defined as:
\begin{equation}
    \nabla_\theta\mathcal{L}_{VSD}(\theta)\approx\E_{t,\epsilon,c}[w(t)(\epsilon_{pretrain}(x_t,t,y)-\epsilon_\phi(x_t,t,c,y))\frac{\partial x}{\partial\theta}]
\end{equation}

\begin{figure}
\centering
\includegraphics[width=\linewidth]{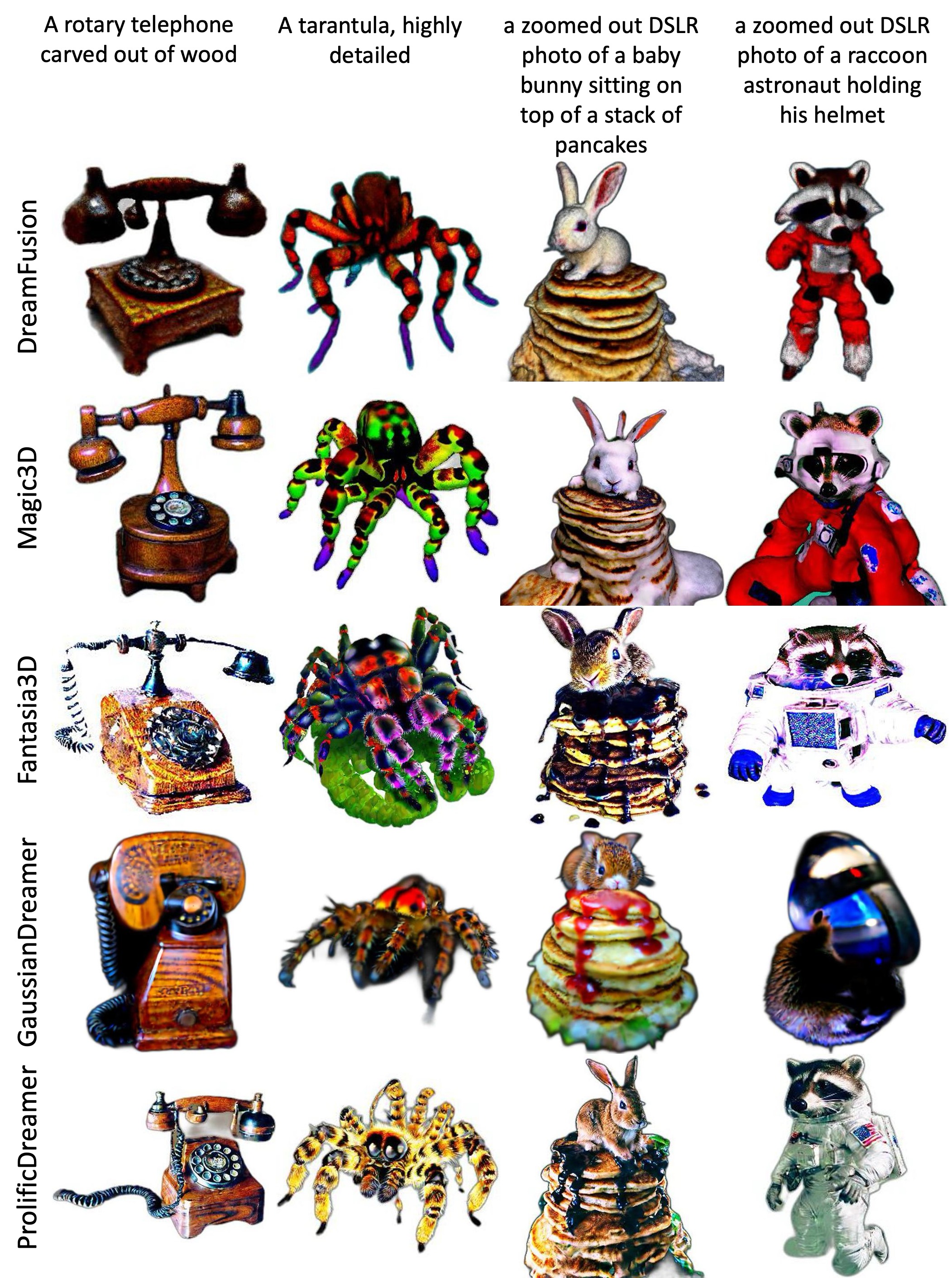}
\caption{
\edited{
Qualitative comparison of DreamFusion~\citep{poole2022dreamfusion}, Magic3D~\citep{lin2023magic3d}, Fantasia3D~\citep{chen2023fantasia3d}, GaussianDreamer~\citep{yi2023gaussiandreamer}, and DreamFusion~\cite{poole2022dreamfusion}.
Results generated by running authors' implementation for GaussianDreamer and threestudio implementation of each method for the others.
This line of work has demonstrated steady progress in improving the quality and coherence of generated 3D objects.
}
}
\label{fig:No3DRep}
\end{figure}

The term $\epsilon_\phi$ represents a LoRA~\citep{hu2021lora} parameterization of the pretrained diffusion network, which also incorporates the camera condition $c$ for rendering images from the NeRFs.
The LoRA network is optimized using the standard diffusion training loss, alternating with the VSD loss.
Though optimizing multiple particles (NeRFs) can be resource-intensive, their model consistently delivers superior results even with a single particle.
Compared to the SDS loss, the produced objects exhibit markedly enhanced quality, as shown in \Cref{fig:No3DRep}.
This approach also enables training at lower guidance weights (7.5 versus 100 in DreamFusion), which helps to produce textures that are more realistic and less cartoonish.
Training is in a two-stage manner with improved hyperparameters similar to Magic3D~\cite{lin2023magic3d} which we discuss in the next section.

\subsubsection{3D Representation Improvements}
\label{sec:5-representation-improvements}

Training of NeRFs is often hampered by the substantial memory requirements of volumetric rendering, which typically restrict methods to lower resolutions.
\edited{Here, we discuss mesh-based representations, such as DMTet, that employ rasterization.
This is a more memory-efficient approach enabling training at higher resolutions, and generation of sharper textures and finer geometric details.
Recently, Gaussian Splatting (GS)~\citep{kerbl20233d} has gained traction due to efficient representation and faster rendering compared to NeRFs.
}

\mypara{Magic3D.}
The Magic3D~\citep{lin2023magic3d} approach addresses the slow training times in DreamFusion and lack of fine details in the generated objects.
These issues stem from the high memory usage of the NeRF model.
The authors present a two-stage coarse-to-fine training pipeline aimed at crafting high-fidelity 3D models.
In the coarse stage they train a Instant-NGP~\citep{mueller2022instant} model at a lower resolution of $64\times64$ under the guidance of the T2I model eDiff-I~\citep{balaji2022ediffi}.
Subsequently, in the fine stage they convert the NeRF model into a DMTet~\citep{gao2020learning, shen2021dmtet} representation.
Training at this stage is done at a higher resolutions of $512\times512$ with Stable Diffusion as guidance.
Thanks to the efficiency of rendering through rasterization rather than volume rendering, this approach considerably reduces the memory footprint.
The resulting 3D meshes extracted from DMTet surpass the quality of those produced by DreamFusion with a substantial $61.7\%$ of human raters preferring the former in a user study.

\mypara{TextMesh~\citep{tsalicoglou2023textmesh}.}
In concurrent work to Magic3D, TextMesh similarly adopts a two-stage approach to avoid over-saturated textures in objects produced from DreamFusion.
In the first stage they train a NeRF model with a SDF backbone for densities.
This makes it easier to extract a mesh after this stage using Marching Cubes~\citep{lorensen1998marching}.
They then render depth from four views of the mesh (top, front and sides) which are tiled and fed through Stable Diffusion with depth conditioning to obtain textured images.
They then fine-tune the mesh textures with photometric loss from the textured images as well as the SDS loss with smaller guidance weights of 7.5 to produce objects that have more photorealistic textures.

\mypara{Fantasia3D~\citep{chen2023fantasia3d}.}
The Magic3D authors stated that DMTet training from random initialization leads to poor results.
This could be due to the more discrete nature of DMTet, whereas NeRFs are more continuous in the sense that there are multiple points along the ray where gradients can propagate.
Fantasia3D shows that it is possible to train from scratch with DMTet alone and achieve high quality geometry as well as textures.
To do this they disentangle the training into separate stages of geometry and appearance modeling with the SDS loss.
Their DMTet geometry model consists of an MLP parameterized by $\Psi$ to predict SDF values and deformations.
The network is initialized with a 3D ellipsoid in the beginning of training.
In the geometry modeling stage they propose a novel mask and normal augmentation technique in the early phase of training directly in the latent space of Stable Diffusion and then encode the normal directly in RGB space for the rest of the geometry training with SDS guidance.
Then, in the appearance training stage they fix the geometry while learning the diffuse $k_d$, metallic $k_{rm}$ and normal variation $kn$ maps for the 3D model using a Physically Based Rendering (PBR) model where another MLP $\Gamma$ is used to predict the different terms.
They show that disentangling the learning of two components results in better geometry as well as better textures due to the PBR rendering model.

\mypara{Gaussian Splatting Methods.}
\edited{
Recent work leverages 3D Gaussians and splatting to efficiently represent complex 3D scenes.
DreamGaussian~\citep{tang2023dreamgaussian} samples points within a sphere and then optimizes the 3D Gaussians with the SDS loss, periodically densifying points to add detail.
GSGEN~\citep{chen2023text} instead initializes using a point cloud generated from Point-E~\cite{nichol2022point}.
The 3D Gaussians are optimized not only with SDS from the rendered images, but a 3D SDS loss with positions of the Gaussian points using Point-E.
They then densify and prune the Gaussians to refine appearance with additional regularization losses along with the image SDS loss.
GaussianDreamer~\citep{yi2023gaussiandreamer} similarly initializes using points from 3D diffusion models such as Shap-E~\citep{jun2023shap}.
These 3D priors used for initialization help generate more consistent shapes.
DreamGaussian and GaussianDreamer boast training speeds of 2 and 15 minutes on a single GPU.
}

\subsubsection{Janus Problem}
\label{sec:5-janus-problem}

\begin{figure}
\includegraphics[width=\linewidth]{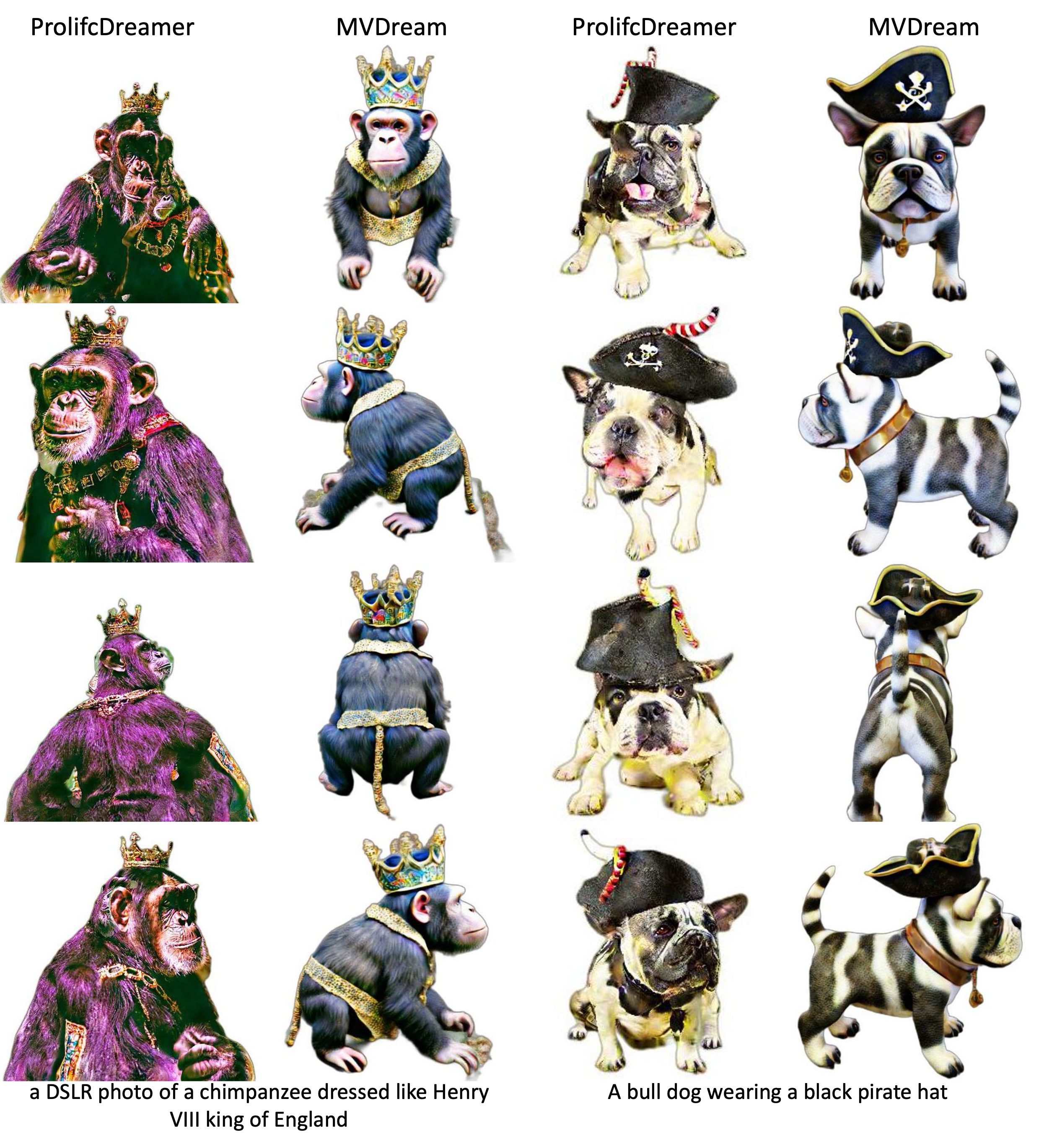}
\caption{
\edited{Qualitative comparison of ProlificDreamer~\citep{wang2023prolificdreamer} vs MVDream~\citep{shi2023mvdream}.
Results generated using MVDream authors' implementation and threestudio ProlificDreamer implementation.
The MVDream approach mitigates the `Janus problem' where the 3D model has multiple frontal faces (each row shows four views around an output 3D model).}
}
\label{fig:MVDream}
\end{figure}

A prevalent issue in text-to-3D shape generation is known as the ``Janus problem''.
This is characterized by the generation of 3D models with multiple faces that deviate from realistic object structure.
For instance, when attempting to create a 3D representation of a dog, certain perspectives may erroneously display multiple frontal faces, as illustrated in the first two rows of \Cref{fig:MVDream}.
This anomaly is likely a consequence of dataset bias in the training of T2I models, with a disproportionate number of images presenting objects from a frontal perspective.
DreamFusion attempted to side step this by introducing view-dependent prompts.
For example, they suggested the use of specific prompts for different azimuth angles: `front view + {prompt}' for angles ranging from $0^\circ$ to $90^\circ$, `back view + {prompt}' for $180^\circ$ to $270^\circ$, and `side view + {prompt}' for the remaining angles, with the addition of `top view' or `bottom view' contingent upon defined elevation thresholds.
Despite these measures offering some improvement, they are not a panacea, as the Janus problem persists across various prompts.

\mypara{Mitigation.}
Some methods have been proposed to side step this issue and can be generally incorporated into SDS-like loss training frameworks.
\citet{hong2023debiasing} propose two methods for debiasing the training process by clipping gradients of the SDS loss, and by removing words that may be in conflict with the viewing angle from the prompts to debias them.
\citet{seo2023let} utilizes 3D geometry priors of shapes generated from models like Point-E~\citep{nichol2022point} to help maintain 3D consistency.
\edited{Initializing 3D Gaussians with points generated by Point-E in GSGEN~\citep{chen2023text} and GaussianDreamer~\citep{yi2023gaussiandreamer} also have the same effect.}
Perp-Neg~\citep{armandpour2023re} proposes to use negative prompts in the T2I diffusion model framework to discourage inconsistent view prompts with respect to the sampled view.
While these methods help mitigate the Janus Problem, they do not solve the underlying bias within T2I models.
\edited{In the following section we discuss methods such as MVDream~\citep{shi2023mvdream} and SweetDreamer~\citep{li2023sweetdreamer} which fine-tune the guidance models to be 3D--aware, directly addressing this issue.}

\subsubsection{Generative Models with SDS Loss}
\label{sec:5-generative-model}

\mypara{ATT3D~\citep{lorraine2023att3d}.}
The ATT3D approach addresses a big problem with the other works discussed in this section: requiring per-prompt training for every object.
ATT3D trains a unified model using the SDS loss with multiple prompts at once.
To do this they add an additional mapping network that takes the text prompt as input and is used to modulate spatial grid features in the Instant-NGP model to generate multiple different objects.
During training they also amortize over text by interpolating over text embeddings, helping to smooth interpolations between different text prompts.
The ATT3D authors show that their approach converges with better efficiency (fewer rendered frames per prompt) and achieves results comparable to the single-prompt training of DreamFusion.
They also show some generalization capabilities with unseen prompts that are compositing new prompts with text in the training prompts.

\edited{At the time of writing another approach~\citep{li2023instant3dSDS} has been proposed to tackle the same issue.
Utilizing a decoder network to generate triplane NeRF representations from text, they also experiment with different ways for effectively injecting the text information.
The model is trained with multi-prompt SDS and CLIP losses with Perp-Neg~\citep{armandpour2023re} to mitigate the Janus problem.
Their method is able to generate higher quality 3D objects compared to ATT3D with better text and 3D alignment.}

\subsubsection{Discussion}

In this section, we first discussed losses for distilling 2D priors learned by T2I models for generating 3D objects.
This includes DreamFusion~\citep{poole2022dreamfusion} and Score Jacobian Chaining (SJC)~\citep{wang2023score} as well as ProlificDreamer~\citep{wang2023prolificdreamer} which improve upon the formulation.
We then introduced works that improve upon the quality of the models with more efficient representations such as DMTet.
The strong priors learned in T2I models demonstrate superior generation quality compared to models in the previous section utilizing CLIP guidance.
\edited{This area is increasingly popular with several followup works analyzing how to improve upon the distillation loss~\citep{liang2023luciddreamer, katzir2023noise, wang2023steindreamer, yu2023text, wang2023taming, tang2023stable, zhou2023dreampropeller, pan2023enhancing, wu2024consistent3d}.}

Compared to works utilizing paired text and 3D or just 3D data, the methods in this family are able to generate objects that span a broader domain and in general can generate objects from more complex text prompts.
However, a drawback of this class of methods is having to train a new 3D model for every text prompt (a limitation that is also present in the CLIP-guided methods).
For higher quality objects, the training time can take up to hours for some methods.
This is in comparison to only having to run a much faster inference for the paired or unpaired 3D data methods.
\edited{ATT3D~\citep{lorraine2023att3d} and Instant3D~\citep{li2023instant3dSDS} address this issue by applying the SDS loss in a multi-prompt generative setting. 
These methods show initial success with better training efficiency compared to per-prompt optimization. 
However, the models are trained on a relatively small amount of prompts and whether this training strategy generalizes to a long-tailed distribution of real objects remains to be seen.
}

\edited{The approaches in this section generate impressive results leveraging 2D priors from T2I models, but the 3D objects are plagued by the Janus problem.
In the next section, we discuss methods that fine-tune T2I models with 3D priors to directly address this issue. %
These methods enable generation of objects within seconds, creating optimism that with continued refinement they may supersede the slower SDS loss optimization approaches.
}

\begin{table*}
\resizebox{\linewidth}{!}
{
\begin{tabular}{@{} l ccrcccccr @{}}
\toprule
Method & InCam & Input & Output Views & Image Res & Multi-View Comm & To 3D & Dataset & Stable Diff & GPU Days \\
\midrule
MVDream~\citep{shi2023mvdream} & abs & Text & 4 RGB & $256^2$ & 3D Attn & SDS & OV + LAION & v2.1 & 3 Days 32 A100 \\
SweetDreamer~\citep{li2023sweetdreamer} & abs & Text & 1 CCM & $64^2$ & -- & SDS & OV filt. (270K) & v2.1 & -- \\
\multirow{2}{*}{Direct2.5~\citep{lu2023direct2}} & \multirow{2}{*}{abs} & Text & 4 Normal & \multirow{2}{*}{$256^2$} & \multirow{2}{*}{3D Attn} & \multirow{2}{*}{Recon} & OV filt.+COYO-400M filt. (500K, 65M) & \multirow{2}{*}{v2.1} & 80 Hours 32 A100 (80G)\\
& & Image (Normal) & 4 RGB & & & & OV filt. (10K) & & 20 Hours 32 A100 (80G) \\
Instant3D~\cite{li2023instant3d} & -- & Text & 1 $2\times2$-tiled RGB & $1024^2$ & -- & LRM* & OV filt. (10K) & SDXL & 3 Hours 32 A100\\
UniDream~\citep{liu2023unidream} & abs & Text & 4 Albedo + Normal & $256^2$ & 3D + Cross Dom Attn & TRM + SDS & OV filt. (300k) + LAION & -- & 19 Hours 32 A800 \\
Zero-1-to-3~\citep{liu2023zero} & rel & Image & 1 RGB & $256^2$ & -- & SJC & OV & IV & 7 Days 8 A100 (80G) \\
SyncDreamer~\citep{liu2023syncdreamer} & rel & Image & 16 RGB & $256^2$ & Cost Vol + Depth Attn & Recon & OV & Zero-1-to-3 & 4 Days 8 A100 (40G) \\
Zero123++~\citep{shi2023zero123++} & -- & Image & 1 $3\times2$-tiled RGB & $960\times640$ & -- & -- & OV & v2 & -- \\
Wonder3D~\citep{long2023wonder3d} & rel & Image & 6 RGB + Normal & $256^2$ & 3D + Cross Dom Attn & Recon & OV (LVIS) & IV & 3 Days 8 A800 \\
\bottomrule
\end{tabular}
}
\caption{\edited{Summary of methods fine-tuning T2I/I2I diffusion models to be 3D aware.  
The `InCam' column indicates whether \textbf{abs}olute or \textbf{rel}ative camera parameters are used to condition the model where \textbf{rel} indicates camera parameters relative to the input reference image (vs absolute coordinates for \text{abs}).
The `Input' column indicates text or image conditioning for generation.
The `Output Views' column shows how many views are generated as the output of the fine-tuned model along with the type of image (CCM, RGB, Normal).
`Image Res', 'Multi-View Comm' indicate the output image resolution of the model and mechanism for communication between multi-views. 
In `To3D' we indicate how the 3D object is obtained either through distillation (SDS, SJC), using images for reconstruction (i.e. photometric losses) or image to 3D models (TRM, LRM*).
LRM* here indicates a large reconstruction model modified to accept multi-views as condition.
`Dataset' and `Stable Diff' indicate the dataset used to fine-tune the model and which Stable Diffusion version the models are based on.
Most of the models use some version of Objaverse (OV), with filtering often applied to discard low-quality assets. 
Note that `--' in UniDream indicates that the specific model used for fine-tuning is not mentioned in their paper.
`GPU Days' shows the time and GPU resource used to train the models.}}
\label{tab:methods-Hybrid3D-T2I}
\end{table*}

\section{Hybrid3D}
\label{sec:hybrid3d}

\begin{figure}
\includegraphics[width=\linewidth]{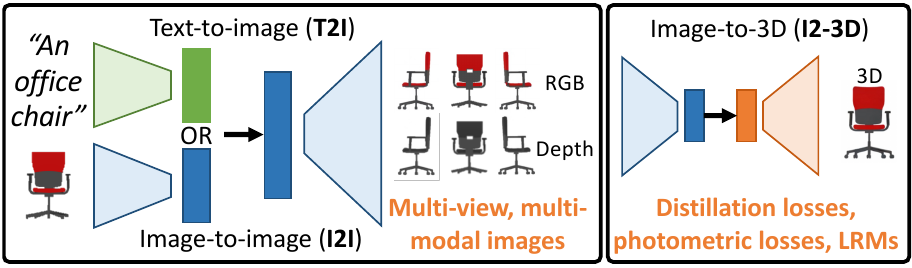}
\caption{
\edited{
Illustration of recent methods that rely on a `Hybrid3D' pipeline of: i) making text-to-image (T2I) or image-to-image (I2I) models 3D--aware through multi-view and multi-modal images and corresponding camera viewpoint information; and ii) improving image-to-3D models using distillation losses, photometric losses and Large Reconstruction Models (LRMs).
}
}
\label{fig:overview-hyrbrid3d}
\end{figure}

\edited{Here, we discuss methods that share a common theme of using images as a bridge between text and 3D, in particular to enforce 3D consistency.
These methods combine text-to-image (T2I) and then image-to-3D (I2-3D) in a pipeline.
This strategy allows use of pretrained T2I models and I2-3D models.
However, two issues need to be addressed:
i) domain gap between output images from pretrained T2I model and input images to pretrained I2-3D model; and
ii) how to incorporate information for improved conditioning of the output 3D model.
The first issue can be addressed by fine-tuning the T2I model with additional data.
The second issue, has been addressed using \emph{3D--aware} T2I models conditioned on camera pose information to encourage consistent 3D shape generation via SDS or photometric losses, or generating multi-view images that are then leveraged for 3D shape generation.
\Cref{fig:overview-hyrbrid3d} provides an overview illustration.
}

\edited{We first describe Point-E~\citep{nichol2022point} as it is a straightforward text-to-image-to-3D pipeline and a good basis for discussing the two-stage pipeline of T2I then I2-3D.
We then focus on each stage:
1) fine-tuning T2I models to be 3D--aware~\citep{shi2023mvdream, li2023sweetdreamer, lu2023direct2, liu2023unidream, liu2023zero, liu2023syncdreamer, shi2023zero123++, long2023wonder3d}; and
2) training I2-3D models to generate 3D objects from single or multi-view images~\citep{nichol2022point, hong2023lrm, li2023instant3d, liu2023one, liu2023oneplusplus}.
We also discuss methods focusing on both stages~\citep{liu2023unidream, li2023instant3d, liu2023one, liu2023oneplusplus}.
Note that we focus on methods that use text as the main input, but briefly touch on methods with image-based input as they can be used for text-to-3D generation in a pipeline similar to Point-E (text-to-image-to-3D).
}

\mypara{Point-E~\cite{nichol2022point}.}
\edited{Point-E adopts a direct approach using text-guided diffusion to first generate an image from text.
The image serves as a condition for a point cloud diffusion model to do single-view 3D reconstruction.
The Point-E approach first refines GLIDE~\cite{nichol2021glide} to generate images similar to synthetically rendered images.
A special token is added to the text prompt to indicate that this is a rendered image so that at inference time the token can signal that the model should generate images similar to rendered ones.
Subsequently, a point cloud diffusion model is trained conditioned on images from the initial stage.
The model is conditioned by leveraging the entire token sequence of the CLIP image embedding from the generated image.
Due to its straightforward combination of the two T2I and I2-3D stages, Point-E serves as a good canonical example of this strategy.
We next discuss work focusing on improving each of these two stages.
}

\subsection{3D--aware T2I}
\edited{
Work focusing on adapting T2I models to be 3D aware uses camera pose information or generates images from different views.
These methods may leverage multi-view images and multi-modal images.
The former involves mechanisms for allowing communication between multi-views during generation, while the latter uses images capturing information beyond RGB (e.g., normal, albedo, depth).
Note that prior work has called the latter multi-modal image inputs `cross domain'~\citep{long2023wonder3d}.
This additional information enables establishing correspondences between views and modalities, leading to improved 3D awareness.
Concretely, multi-view methods adapt the T2I model to generate multiple views at once and modify the self attention layer of diffusion models to enable 3D attention across views~\citep{shi2023mvdream, lu2023direct2, liu2023unidream, long2023wonder3d}, tile multiple views in a single image~\citep{shi2023zero123++, li2023instant3d}, or use a unified 3D representation during generation~\citep{liu2023syncdreamer}.
Multi-modal image methods generate several image modalities at once by leveraging cross domain attention~\citep{liu2023unidream, long2023wonder3d}.
We organize these methods into text conditioning and image conditioning.
}

\mypara{Text conditioning.}
\edited{Several recent methods fine-tune existing T2I models to be 3D aware and take text as the input condition. MVDream~\citep{shi2023mvdream} and SweetDreamer~\citep{li2023sweetdreamer} aim to solve the Janus problem and improve object coherency by swapping the guidance models with T2I models fine-tuned with 3D priors.
Direct2.5~\citep{lu2023direct2} devises a optimization scheme using the sparse multi-view geometric and color images generated from the fine-tuned T2I models to enable fast generation of objects in 10 seconds.
UniDream~\citep{liu2023unidream} designs a pipeline that is able to generate re-lightable objects with PBR materials.}

\mypara{Image conditioning.}
\edited{We briefly mention image-to-image (I2I) methods that generate consistent novel views of a 3D object given a reference image as input, but do not go into  detail as they are beyond the focus of this survey.
A popular choice for fine-tuning is Image Variations (IV)~\citep{sd_image_variations_lambdalabs} which fine-tunes Stable Diffusion to condition on images as input instead of text.
Zero-1-to-3~\citep{liu2023zero} was one of the first to leverage priors in existing diffusion models for 3D reconstruction. 
They fine-tune the model to generate novel views of the input reference view with relative camera parameters of the target view as a condition, and apply the SJC loss for 3D generation.
SyncDreamer~\citep{liu2023syncdreamer} builds on Zero-1-to-3 and adds a unified cost volume to generate multiple views at once, resulting in better consistency between views.
Zero123++~\citep{shi2023zero123++} proposes several training schemes to improve the stability of the fine-tuning process and output quality.
Here, multiple views are tiled into one image to enable the generation of multiple frames at once.
Wonder3D~\cite{long2023wonder3d} proposed a domain attention to allow for the generation of multi-view RGB and normal images at once.
With the additional geometric information from the normal maps, they train a NeuS~\citep{wang2021neus} model to reconstruct meshes from the sparse views in 2 to 3 minutes.}

\subsection{Image-to-3D Models}

\edited{Work targeting this stage uses three strategies to generate 3D objects from images:
1) distillation losses like SDS;
2) image reconstruction from sparse views with photometric losses; and
3) separate model generating 3D objects from conditioning images.
}
\edited{
\Cref{tab:methods-Hybrid3D-T2I} organizes work using these three strategies to obtain 3D models from fine-tuned T2I diffusion models.
}

\mypara{Distillation losses.}
\edited{
Methods using distillation losses (like in No3D) produce objects of higher quality, with the optimization gradually transforming a 3D representation over many iterations with different camera views sampled, allowing for more intricate details to be generated. 
However, the main limitation is the high generation time per 3D object, similarly to No3D approaches.
}

\mypara{Image reconstruction from sparse views.}
\edited{
Methods that optimize the 3D objects directly to match the images via reconstruction methods are much faster (e.g., 10 seconds in Direct2.5).
Typically this speed is obtained by having a limited number of sparse views.
The main drawback is that the sparse viewpoints make it hard to generate more complex objects due to occlusions.
Also, eliminating inconsistencies between views is hard, which can create artifacts in the extracted 3D object.
}

\mypara{Trained image to 3D models.}
\edited{
Due to increasing 3D data, there is growing interest in training a separate image to 3D model.
This requires running inference on the model which is usually much faster (e.g., 20 seconds in Instant3D) than optimization.
However, these models need larger capacities to get high quality 3D objects leading to higher resource requirements.
Also, depending on the dataset used to train the model, generalizability may be limited.
A popular alternative to diffusion models are transformer-based methods called Large Reconstruction Models (LRM)~\citep{hong2023lrm}.
In LRM, a transformer predicts triplane NeRF representations with conditioning injected through cross attention from a single input image.
Both Instant3D~\cite{li2023instant3d} and UniDream~\citep{liu2023unidream} use LRM and modify it to accept multiple views at once.
Their pipeline is similar to Point-E, except a text to multi-view image network is fine-tuned.
One-2-3-45~\citep{liu2023one} and One-2-3-45++~\citep{liu2023oneplusplus} leverage SparseNeuS~\citep{long2022sparseneus} and LAS-Diffusion~\citep{zheng2023locally} respectively as backbones for multi-view image to 3D generation.
}

\subsection{Discussion}

\edited{Some challenges in using these `Hybrid3D' methods revolve around what datasets to use for fine-tuning and what camera views to select for rendering.
One problem with using Objaverse~\citep{deitke2023objaverse,deitke2023objaversexl} the largest available synthetic 3D object dataset is that object textures appear toy-like and flat making generated objects less photorealistic.
SweetDreamer addresses this by using normal map images and an unmodified T2I model for texture optimization.
UniDream generates albedo and normal maps, and then generating 3D objects with another reconstruction model with a Stable Diffusion model used to optimize for photorealistic PBR materials.
The choice of camera viewpoints used to fine-tune the model is also important.
For example, MVDream samples cameras from the upper hemisphere of the object meaning artifacts may be visible when looking at the bottom of the object.
Methods utilizing direct reconstruction may choose fixed viewpoints (e.g., Direct2.5).
However, depending on the type of object being generated important details may be occluded.}

\edited{
Looking at the dataset column in \Cref{tab:methods-Hybrid3D-T2I} we see different strategies for filtering datasets.
Instant3D adopts an interesting strategy by training an SVM on manually labeled examples to filter the dataset.
In general, Objaverse is noisy and can hurt model performance if used in entirety.
However, how to best filter the dataset while not hurting the generalizability of models is not well studied.
}

\edited{The GPU days column in \Cref{tab:methods-Hybrid3D-T2I} shows these models are expensive to train.
Diffusion models especially require large batch sizes for training. Naive scaling to generate more dense views leads to prohibitive GPU requirements. It would be interesting to explore how large the base diffusion models need to be to learn generation of multi-view images. For example, is it possible to use smaller distilled models like BK-SDM~\citep{kim2023architectural} to fine-tune these methods?}

\edited{Another promising future direction is exhibited by DMV3D~\citep{xu2023dmv3d}, where the model is designed to generate consistent multi-view images. The multi-view diffusion model uses a transformer model from LRM to predict an internal NeRF during the denoising process and encourage consistency between the multi-view images being denoised.}

\section{Generating multi-object scenes with diffusion guidance}
\label{sec:method-scene}

\begin{figure}
\centering
\includegraphics[width=\linewidth]{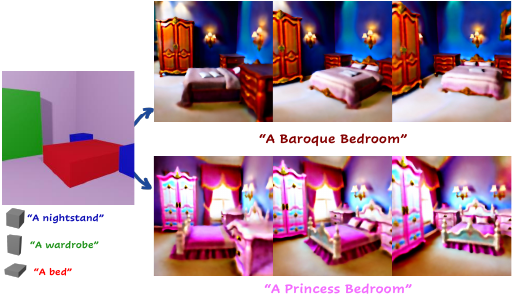}
\caption{Example input and output from Set-the-Scene~\citep{cohen2023set}.
On the left, bounding boxes indicate the location of each object a well as their local accompanying text.
On the right the outputs resulting from different global text prompts are shown.
Visuals reproduced from \citet{cohen2023set}.}
\label{fig:SetTheScene}
\end{figure}

The work discussed thus far in this survey mainly focuses on generating a single object from an input text prompt.
While it is possible to use more complex prompts describing a multi-object scene, the methods we described often fail to generate a coherent scene composition incorporating multiple objects. 

There is a long line of prior work on 3D scene layout generation, which typically assumes the presence of a 3D shape database from which objects are retrieved and composited into a scene layout.
Some seminal work in this vein, and guided by input text descriptions was done in \citet{chang2014learning}.
An excellent survey covering the earlier 3D scene layout work in detail is provided by \citet{chaudhuri2020learning}.
In this survey, we focus on generation of 3D scenes built on top of the methodology presented thus far, which offers the advantage of generating scene layouts composed of objects each of which is also generated conditioned on the input description. There has been limited work on this latter strategy to text-to-3D scene.

In contrast to the object-centric approaches discussed earlier in the survey, it is also possible to generate an entire scene without explicitly modeling individual objects.
These methods typically rely on text-to-image (T2I) models to generate partial views of the scene based on the text, and use outpainting methods to generate additional views.
These views are then fused together by using depth prediction.
Here, we describe these different lines of work that all aim to generate more complex 3D scene-level outputs. 

\subsection{Compositional Generation}
\label{sec:method-scene-composition}

The papers in this section aim to generate a scene consisting of multiple objects, \edited{mostly} assuming that a layout of objects is given as a condition in the form of bounding boxes and the accompanying text prompts.
An example of this from Set-the-Scene~\citep{cohen2023set} is shown in \Cref{fig:SetTheScene}.  \edited{More recent works~\cite{vilesov2023cg3d,gao2023graphdreamer,zhang2023scenewiz3d} drop the assumption that the layout is given and attempt to both predict the layout (scale, position, and rotation) for each object as well as generate the individual objects.}

\mypara{Set-the-Scene.}
\citet{cohen2023set} composes individual Latent-NeRF models for each of the object bounding boxes in the input layout condition.
They propose a interleaved training scheme where individual NeRFs as well as the entire scene where all the NeRFs are rendered is trained with the SDS loss alternately. %
This allows for the optimization to focus on local objects as well as making sure that individual objects synergies with each other in the global frame.
Compared to their baseline Latent-NeRF which fails to generate complex scenes their method is able to generate coherent scenes with individual objects.
As the objects are all individual NeRFs, they can be rearranged and placed into a scene layout through appropriate spatial transformations.

\mypara{CompoNeRF.}
\citet{lin2023componerf} similarly uses multiple Latent-NeRF models for each object bounding box.
However, instead of simply adding the density and color components of each NeRF during volumetric rendering they add additional global and local MLPs to further refine density and colors in the rendered global frame.
This helps to make the objects more consistent with each other in the scene.
The individual and global SDS losses are added and optimized in conjunction.

\mypara{Comp3D.}
\citet{po2023compositional} offers a different approach instead of training multiple NeRF models.
They calculate the classifier-free guidance for each individual prompt of the objects in the scene.
Then, using the rendered bounding boxes as segmentation masks the gradients for each region are masked and combined to be optimized.
This has the benefits of less memory usage compared to the above methods, as only one NeRF model is being trained.
This model is based on the Score Jacobian Chaining (SJC) strategy described in \Cref{sec:method-unsupervised-diffusion}.

\mypara{Layout Generation.}
\edited{The above methods assume that bounding box layouts are given.
Recent work addresses layout generation together with object generation.  
CG3D~\cite{vilesov2023cg3d} uses a probabilistic graphical model (PGM) to sample objects and estimate their interactions with each other.  The text description is converted to a scene-graph which is used to instantiate the PGM.  The scene is generated by optimizing SDS losses to generate objects (each represented as a set of 3D Gaussian) and a combination of SDS loss and physical constraints (e.g. gravity and contact) loss for object interactions (e.g. the placement of objects relative to each other).  The integration of phyical constraint losses allows for plausible generation of objects that are \emph{on} or \emph{in} another object. 
Another promising direction is the use of LLMs for layout generation.
GraphDreamer~\citep{gao2023graphdreamer} generates a scene graph using LLMs such as ChatGPT.
The scene is then optimized using several SDS losses on individual objects, the entire scene and objects with relations (edge pairs in the scene graph).
SceneWiz3D~\citep{zhang2023scenewiz3d} similarly uses an LLM to generate objects of interest in a text prompt and then uses an off-the-shelf text-to-3D model to generate initial 3D objects.
A particle swarm optimization algorithm with CLIP similarity then optimizes the object locations, followed by optimization with VSD losses for the objects and the environment.}

\subsection{RGBD Fusion for Scenes}
\label{sec:method-scene-fusion}

While distillation losses like SDS have been successful in object-centric generation, crafting detailed scenes with complex object compositions remains challenging.
The research discussed in this section harnesses the priors learned by T2I models for scene generation.
In contrast to prior work relying on the SDS loss to slowly distill knowledge from T2I models.
They integrate images predicted from T2I models and a depth estimation network to perform RGBD fusion, thus generating a 3D mesh or NeRF representation for the entire scene.

\mypara{SceneScape~\citep{fridman2023scenescape}.}
The SceneScape approach is the first to tackle the task of perpetual view generation given text as input.
It aims to create a video of a consistent 3D scene using a provided text and camera path.
To achieve this, SceneScape employs two pretrained models: Stable Diffusion (SD) inpainting and a depth estimation model~\citep{ranftl2021vision}.
The process starts by generating an initial image from the SD model using the text prompt, complemented by depth predictions.
From this, an initial mesh is established.
For each subsequent frame, the mesh is rendered based on the updated camera position, leading to a frame with gaps.
These gaps are filled using the inpainting model, while the depth estimation model offers the necessary depth predictions.
The updated image and depth data is then used to refine the mesh for subsequent frames.
After each frame, the decoder of both the inpainting and depth models is fine tuned and reset for frame consistency.
SceneScape generates videos of diverse scenes with intricate details.
Yet, the method is subject to limitations.
Over time, errors can accumulate and this technique struggles particularly with outdoor scenes.

\mypara{Text2Room~\citep{hollein2023text2room}.}
The Text2Room approach has a comparable strategy to SceneScape, merging texture and geometric data over several frames to create a 3D mesh for room generation from text.
A Stable Diffusion inpainting model and a depth prediction mechanism iteratively refine frames and determining depths to continually enhance the mesh representation.
To ensure depth consistency across frames, a specialized depth inpainting model is deployed.
Moreover, to fine-tune alignment the scale and shift parameters of the predicted depth map are optimized, ensuring a more accurate match with the known depth derived from the mesh.
Using the pixel depth values a point cloud is constructed where every four neighboring pixels are linked to produce two triangles.
Further filtration removes faces that could lead to visual distortions.
Unlike SceneScape's continuous scene creation, indoor rooms possess complex structures at many spatial scales.
This distinction amplifies the significance of camera trajectory choices, ensuring the generated rooms possess plausible structures, layouts, furnishings, and minimizing gaps in the geometry.
Text2Room utilizes a two phase viewpoint selection technique.
Initially, they leverage a set of predefined camera trajectories to establish the primary scene layout and furniture planing.
In the subsequent phase, specific views are chosen to refine and fill in any geometric hole.
The final mesh undergoes Poisson surface reconstruction~\citep{kazhdan2006poisson} to refine its form.
While their approach successfully crafts comprehensive 3D scenes with detailed textures, some outputs may exhibit stretched geometric areas or overly smoothed regions.

\mypara{Text2NeRF~\citep{zhang2023text2nerf}.}
The Text2NeRF approach also aims to solve the task of text to scene generation.
Differing from the mesh-based approach of previous methods, it uses an implicit NeRF network for its 3D representation.
They start by generating an initial image using Stable Diffusion, accompanied by depth predicted from a depth estimation model.
Subsequently, by altering camera positions and projecting the initial image and depth with the corresponding camera parameters, they generate a support set of images and depths.
These will naturally contain gaps.
Both the initial and support are then used to initialize the NeRF model.
Then, a new view is chosen to be rendered via NeRF.
The rendered image and depth containing holes are filled in with Stable Diffusion inpainting, while the depth is estimated using the depth network.
Depth alignment adopts a two stage approach: initially, by calculating the mean scale and distance disparities for alignment, and then by fine-tuning a neural network for nonlinear depth alignment.
The NeRF model is further trained with the updated frame, with the cycle of rendering new views, inpainting and updates repeating.
The main advantage of Text2NeRF lies in its use of an implicit representation such as NeRF, which avoids issues such as geometric stretching seen in Text2Room, notably in outdoor scene settings.

\section{Editing}
\label{sec:method-editing}

An important consideration in the text to 3D shape generation task is allowing the user to apply edit operations on an output 3D shape using text instructions.
This functionality is clearly valuable in practical scenarios where the user desires specific changes in the output 3D shape.
Enabling such intuitive editing operations constitutes a significant open research avenue.
Here, we discuss recent work that focuses on text-based 3D editing by optimizing 3D shapes with CLIP (\Cref{sec:shape-editing-clip}) or leveraging text-to-image models for 3D editing (\Cref{sec:scene-editing-t2i}).
As changing the appearance of an object while keeping the geometry the same is an important use case, we also discuss recent work in retexturing based on a text prompt (\Cref{sec:texturing}).

\subsection{Shape Editing with CLIP}
\label{sec:shape-editing-clip}

One way to edit 3D shapes given a text command is to optimize the 3D representation so that the rendered image and text match by using CLIP guidance.

\mypara{CLIP-NeRF~\citep{wang2022clip}.}
CLIP-NeRF employs CLIP's guidance to manipulate 3D objects using text.
Initially, a disentangled conditional NeRF model is trained.
In this model, the input condition is separated into a shape code $z_s$ and an appearance code $z_a$.
Notably, the latter only influences color predictions, while the former is used to deform positional encodings within the NeRF network.
The entire model's optimization is carried out using a GAN loss.

For image or text-driven manipulations two distinct mappers, the shape mapper $M_s$ and the appearance mapper $M_a$, are trained to utilize the CLIP text-image encodings to predict update codes, which are subsequently added into the shape and appearance codes of the conditional NeRF model.
This ensures alignment with the given input text or image description.
The mappers' training is supervised by both the discriminator and CLIP similarity losses.
Crucially, during this phase, only the mappers are trained while all previously trained modules remain frozen.

\mypara{Text2Mesh~\citep{michel2022text2mesh}.}
Given a text prompt, Text2Mesh aims to stylize a 3D object mesh by adding 3D geometric detail and color.
To do so, Text2Mesh takes the given shapes and optimizes displacement vectors for each mesh vertex and its color attribute.
The training predominantly relies on CLIP similarity, with some regularization strategies to avert implausible geometry.

\subsection{Scene Editing with Text-to-image Models}
\label{sec:scene-editing-t2i}

Recent advances in text-to-image (T2I) models can be leveraged for editing 3D objects or scenes.
The T2I models can be used to provide edited versions of rendered images that are then reincorporated into the 3D representation~\cite{haque2023instruct,kamata2023instruct}.
Here we briefly describe a few works that use T2I models for editing of 3D objects or scenes.

\mypara{SKED.}
\citet{mikaeili2023sked} provides a method for sketch and text guided editing of an object.
The input comprises of multiple sketch-views consisting of closed shapes indicating regions for edit as well as the complementing text prompt.
The base object is represented as an Instant-NGP NeRF model.
The overall optimization is guided by the SDS loss to ensure the object is aligned with respect to the text prompt.
However, to preserve the identity of the base object they propose two losses to regularize training.
A preservation loss that penalizes density and color changes with respect to the original object especially for points distant from the designated edit areas, and a silhouette loss to make sure that the edit regions are being filled.
The results show that this method can perform localized edits of 3D objects while preserving the original identity of the object.

\mypara{Vox-E.}
\citet{sella2023vox} aims to perform local or global edits of a 3D object given a text prompt.
They start off with a voxel grid NeRF model (ReLU Fields) initialized with the original object. They then train the model with the SDS loss and a volumetric regularization using a correlation loss between densities of the original and edited objects. While this objective produces an output aligned with the text prompt and structure that mostly preserves the original identity. It may be desirable for local edits to keep other regions unchanged. To address this, a spatial refinement method has been introduced. This method segments the locally-edited region from the modified object and merges it with the original. The process utilizes the cross-attention modules in the T2I model to generate attention maps with respect to the the text indicating the local edit as well as the rest of the text. Using the attention map, two additional NeRF models are trained. Subsequently, an energy minimization approach is employed to derive the segmentation mask used for merging. With this their method is able to perform global edits that can change the contents of the entire object, as well as more localized edits that only modify a small region.

\mypara{Instruct-NeRF2NeRF.}
\citet{haque2023instruct} is designed to modify NeRF scenes using textual inputs.
The procedure alternates between updating input images and training the NeRF model.
InstructPix2Pix is used to edit the images based on the provided text. These images include both the multi-view dataset used for NeRF training and images rendered with different camera parameters.
This cycle of updating the dataset and then training the NeRF with the refreshed images is termed 'Iterative Dataset Update' or Iterative DU.
Initially, the edits will be inconsistent, but with continued iterative updates, the NeRF and the rendered images progressively align and converge to achieve a uniformly consistent scene.

\mypara{Instruct 3D-to-3D.}
\citet{kamata2023instruct} similarly uses InstructPix2Pix for guiding the editing process of a NeRF scene with a text prompt.
However, they instead leverage the modified score estimate in InstructPix2Pix to calculate the SDS loss which is used to optimize the scene.
Together with a dynamic scaling scheme, the number of voxels in the DVGO model used for training is gradually reduced than increased again during training to facilitate global edits of the global structure then fine details.

\mypara{Room Dreamer.}
\citet{song2023roomdreamer} transforms the scene texture of an input room scene given a text prompt while improving the geometry.
They start off by rendering a cube map of images, depths and distance maps from the center of the input room scene.
Then leveraging these and the text prompt as input they generate stylized images through a T2I model.
Gaps in the scene not rendered by the cube map are outpainted and filled.
The images are then used to optimize the textures of the original input scene.
To address potential artifacts in the original scene, a geometric loss is incorporated, ensuring that both the rendered depth and the depth predicted by an estimation network remain consistent and smooth.

\subsection{Texturing}
\label{sec:texturing}

One common edit operation on 3D shapes is re-texturing, which can help to create a variety of fine-grained surface appearance for the same shape geometry.
The papers in this section utilize depth conditioned image generation diffusion models for texture generation of 3D objects.

\mypara{TEXTure.}
\citet{richardson2023texture} proposes a method for texture generation of 3D meshes using Stable Diffusion-based depth-to-image and inpainting models.
Their method involves a progressive process of rendering depth images from different views of the 3D object.
Then, they use the depth-to-image diffusion model to generate images based on the text prompt and depth.
The image is written back to a texture atlas of the 3D object with UV mapping calculated by XAtlas.
To ensure local and global consistency of the painted textures from different views, they use another meta-texture map to keep track of which regions of the texture are classified as ``keep'', ``refine'' or ``generate''.
The ``keep'' regions are regions that are fixed during the update process as other views have already generated the texture for the region.
The ``refine'' regions occur when the view used to paint the region is oblique resulting in high possibility of distortions.
The ``generate'' regions have yet to be generated by other views.
Given these segmented regions of the meta-texture, a mask is generated during the sampling steps of the image to keep regions marked as ``keep'' unchanged, as well as allowing for changes in ``refine'' and ``generate'' regions.
The authors also find that alternating between the depth-to-image and inpainting models during the sampling steps help with consistency.
Finally, the generated image is used to update the texture map for regions marked as ``refine'' and ``generate''.
They show that the generated textures can have higher quality as well as consistency compared to prior methods.
Additionally, their method can be utilized for other tasks such as texture transfer, texture from images and texture editing.
However, some drawbacks include global inconsistencies and the fixed viewpoint selection resulting in some geometries not being fully covered by the texture generation process.

\mypara{Text2Tex.}
\citet{chen2023text2tex} is concurrent work to TEXTure and offers a similar strategy for generating textures by progressively painting textures from different views with a depth-to-image model.
First, given a set of fixed viewpoints, they calculate similarity masks which indicates how perpendicular the view direction is to the face normals of the mesh.
Using the similarity masks they segment into generation masks of ``new'', ``update'', ``keep'' and ``ignore'' regions similar to TEXTure.
The ``new'' regions are inpainted from random noise, and the ``update'' regions have a smaller denoising strength.
Other regions are fixed and masked respectively.
Different from TEXTure, Text2Tex adds an additional texture refinement stage with an automatic viewpoint selection scheme.
For the views sampled in this stage, a view heat is constructed with the generation mask indicating areas that may have artifacts in textures.
They then dynamically select views with high view heat and update those using their algorithm.
This helps to eliminate additional distortions in the texture that were not visible from the fixed camera viewpoints.
This method generates high quality textures, with the automatic viewpoint selection helping to eliminate additional artifacts.

\mypara{SceneTex}
\edited{
\citet{chen2023scenetex} proposes an optimization-based approach for generating scene textures. They parameterize the scene appearance with a multi-resolution texture field with cross attention texture decoder. The implicit textures are optimized through a VSD loss and a diffusion model with depth prior as well as the input text. This method generates textures with more detail and consistency with the text compared to prior methods, but comes at the cost of long optimization times per scene.}

\section{Evaluation}
\label{sec:evaluation}

As text-to-3D shape generation methods develop, there will be an increasing need for systematic evaluation to assess the strengths and weaknesses of the different methods.
\edited{Recent work~\cite{wu2024gpt} recognizes the need for automated evaluation of text-to-3D and proposes the use of large vision-language models such as GPT-v4 for automated evaluation.}
In this section, we start by identifying critical dimensions along which text-to-3D shape generation methods should be evaluated (\Cref{sec:eval-desiderata}), and then summarize current evaluation methodologies (\Cref{sec:eval-existing}).

\subsection{Desiderata}
\label{sec:eval-desiderata}

Following work in text-to-image generation, it is critical to evaluate text-to-shape generation along the following dimensions which characterize the output.

\mypara{Quality.}
This axis measures \emph{how good are the generated shape?}
The simplest way to think of this dimension is in terms of how realistic the generated shape appears to a human observer.
With 3D shapes, there are additional properties of quality of the output shape including geometric consistency (does the shape exhibit the Janus problem, are there disconnected parts when there should not be any), mesh quality (is the mesh topology smooth where it should be, and does it have enough geometric detail in complex regions), as well as color and texture quality. 

\mypara{Fidelity.}
This dimension involves measurement of \emph{do the generated shapes match the text prompt?}
In text to 3D shape generation, it is important not only that that the shape is of high quality but that it also matches the text specification.
Characterization of such fidelity measures requires computing the degree to which properties specified by the text are respected in the output 3D shape.

\mypara{Diversity.}
This axis answers \emph{do the generated shapes exhibit variety?}
In other words, another important property for a good generative model is the ability to generate a diverse set of shapes.
Given that a variety of shapes can satisfy the same text description (e.g. there are many chairs that match the description \textit{black chair}), it is important that multiple, diverse shapes can be generated.
However, it is difficult to know the full distribution of the space matching the text description and measuring whether the full set shapes is covered that distribution well is challenging. 

\mypara{Compositionality.}
Metrics measuring this dimension attempt to answer the question \emph{can the method handle text describing different combinations of parts, attributes, and spatial relations?}
Only recently has text-to-2D generation work started to more systematically investigate this aspect~\cite{park2021benchmark}.
Measurement in this dimension requires systematic evaluation of generations for text description that exhibit the compositional nature of language (e.g. \textit{black chair with gray legs} vs \textit{gray chair with black legs}).

\mypara{Speed and efficiency.}
In addition to assessing the above qualities, it is also important to compare the speed and memory resources required.
Methods are typically measured in terms of wall clock time for training, generating a single output (i.e. inference), and the memory needs for either training or generation.
For reference, \Cref{tab:speed-mem-comparison} provides a summary of reported speed and memory consumption for methods in this survey.

\begin{table}[t]
\resizebox{\linewidth}{!}
{
\begin{tabular}{@{} rr rrrr @{}}
\toprule
Type & Method & Train & Per-prompt & Device & Gen time \\
\midrule 
\paired & TITG3SG~\citep{liu2022towards} & yes & no & V100-32G &  2.21s (24.44s) \\
\paired & Shap-E~\citep{jun2023shap} & yes & no & V100 & 13s \\
\rowcolor{verylightgray}
\unpaired & TAPS3D~\citep{wei2023taps3d} & yes & no & V100-32G &  0.05s (7.09s) \\
\midrule
CLIP-Guide & DreamFields~\citep{jain2022zero} & no & yes & TPU cores & 72m \\
CLIP-Guide & PureCLIPNeRF~\citep{lee2022understanding} & no & yes & GTX 2080ti & 20m \\
Diff-Guide & DreamFusion~\citep{poole2022dreamfusion} & no & yes & TPUv4 & 90m \\
Diff-Guide & SJC~\citep{wang2023score} & no & yes & RTX A6000 & 25m \\
Diff-Guide & Prolific Dreamer~\citep{wang2023prolificdreamer} & no & yes & A100 & several h \\
Diff-Guide & Magic3D~\citep{lin2023magic3d} & no & yes & 8x A100 & 40m \\
\edited{Diff-Guide} & \edited{DreamGaussian~\citep{tang2023dreamgaussian}} & \edited{no} & \edited{yes} & \edited{V100} & \edited{2m}\\
\edited{Diff-Guide} & \edited{GSGEN~\citep{chen2023text}} & \edited{no} & \edited{yes} & \edited{4x RTX 3090} & \edited{30m}\\
\edited{Diff-Guide} & \edited{GaussianDreamer~\citep{yi2023gaussiandreamer}} & \edited{no} & \edited{yes} & \edited{RTX 3090} & \edited{15m}\\
Diff-Guide & Fantasia3D~\citep{chen2023fantasia3d} & no & yes & 8x RTX 3090 & 31m\\
Diff-Guide & MVDream~\citep{shi2023mvdream} & no & yes & V100 & 1-1.5h\\
Diff-Guide & SweetDreamer~\citep{li2023sweetdreamer} & no & yes & ?/4x V100 & 20m/1h\\
\midrule
\edited{Hybrid3D} & \edited{Point-E~\citep{nichol2022point}} & \edited{yes} & \edited{no} & \edited{A100} & \edited{25s}\\
\edited{Hybrid3D} & \edited{Direct2.5~\citep{lu2023direct2}} & \edited{yes} & \edited{yes} & \edited{A100} & \edited{10s}\\
\edited{Hybrid3D} & \edited{Instant3D~\citep{li2023instant3d}} & \edited{yes} & \edited{no} & \edited{A100} & \edited{20s}\\
\bottomrule
\end{tabular}
}
\caption{Comparison of reported speed and memory needed for different methods.
Numbers in parentheses indicate type to generate a mesh representation from the base 3D representation (neural fields or voxels for TITG3SG).
Average per-shape inference time estimates are taken fron the respective papers, except for TITG3SG which was from TAPS3D~\cite{wei2023taps3d}.
Point-E~\cite{nichol2022point} gave a range for inference time depending on model size (ranging from 16s to 1.5m for 40M parameter to 1B), we take the 300M parameter condition that Shap-E~\cite{jun2023shap} also uses for comparison.
}
\label{tab:speed-mem-comparison}
\end{table}

\subsection{Existing evaluation}
\label{sec:eval-existing}

Pioneering works such as Text2Shape~\cite{chen2019text2shape} and DreamField~\cite{jain2022zero} not only show that is possible to generate 3D shapes from text, but also propose quantitive evaluation protocols that can be used.
However, many followup works do not perform any quantitive evaluation and provide qualitative examples only.
This is especially true for works that focus on proposing new guidance losses~\citep{wang2023score,wang2023prolificdreamer} and improved 3D representations~\cite{metzer2023latent, chen2023fantasia3d}.
Below we organize the types of evaluation protocols in prior work, and outline directions for more comprehensive evaluation of text-to-3D shape generation.

\mypara{User studies.}
As it is challenging to evaluate the output of a generative model, it is typical to use user studies that compare the output from different systems.
One type of user study is A/B testing where users are asked to compare outputs from two systems.
As the user response can depend on the specific question asked, it is common to pose several different questions to users to extract user judgements corresponding to different dimensions such as quality and fidelity.
For instance, \citet{tsalicoglou2023textmesh} asked users preferences on natural colors, detailed textures, and visually preferred.
However, this principle is not always followed.  Some work, such as \cite{lin2023magic3d}, only ask users to broadly judge the quality of the generated shape (they ask users to select the one that is more realistic).
 
\mypara{Evaluation against ground-truth shape.}
Text2Shape\cite{chen2019text2shape} evaluated generation results using several quantitative metrics that required access to a ground-truth shape: Intersection-over-Union (IoU), Earth Mover's Distance (EMD), and classification accuracy (Class Acc).
The metrics aimed to evaluate the geometric accuracy (IoU) and color (EMD) against a ground truth shape associated with the description, and in general whether the generated shape matches the class (table vs. chair in their case).
These metrics are easy to compute algorithmically and quantify the underlying properties precisely.
However, there is a strong assumption of only one ground truth output which is unrealistic and in tension with the desire for diversity in the output.
In addition to the above, Text2Shape also evaluated their results using a variant of the Inception Score\cite{salimans2016improved} using a shape classifier.
The Inception Score metric combines \emph{quality} (can the classifier identify between the shapes) and \emph{diversity} (do generated shapes exhibit the class distribution).

\mypara{FID.}
Another way to measure the \emph{quality} of a generated shape is to evaluate how natural the rendered images of the generated shapes appear to be to a neural architecture that was trained on image data.  
For instance, \citet{tsalicoglou2023textmesh} evaluated renderings using an FID metric based on CLIP.
Other papers that use this FID metric to evaluate quality include \citet{xu2023dream3d,wei2023taps3d}.

\mypara{Point cloud metrics.}
In addition to evaluating the quality of the renderings, it is also possible to attempt to measure the \emph{quality} of the geometry of the shape using point clouds.
Recently, point-based versions of the inception score~\cite{salimans2016improved} (P-IS) and Frechet Distance~\cite{heusel2017gans} (P-FID) were introduced by \citet{nichol2022point} and used in follow-up work~\cite{zhao2023michelangelo}.
Similar metrics include the Frechet Point Cloud Distance (FDP) which has been used in the shape generation community~\cite{achlioptas2018learning}.
This metric has also been used for evaluating text-to-shape generation~\citep{wei2023taps3d}.

\mypara{Automated pairwise comparison.}
Given a set of shapes (targets and distractors) for a text description, these evaluation protocols train a neural evaluator to select the correct shape.
The neural evaluator can then produce a confidence score for each shape.
This evaluation protocol was introduced by ShapeGlot and trained with the ShapeGlot dataset~\cite{mittal2022autosdf,cheng2023sdfusion}.
This evaluation strategy is used to compare two methods, given a shape generated by method 1 and shape generated by method 2.
If the confidence score for the two methods is within a certain threshold (0.2), then the evaluator cannot determine which of the two classes the shape is from, and is confused. 
The method selected more often by the neural evaluator is said to perform better.

\mypara{Retrieval model R-Precision.}
The CLIP R-Precision metric was introduced by \citet{park2021benchmark} and popularized for evaluation of text-to-3D shape generation in DreamFields~\cite{jain2022zero}.
R-Precision measures the fraction of generated shapes that are retrieved correctly using a retrieval model based on CLIP similarity of rendered views to the text prompt.
DreamFields used a set of 153 text queries and corresponding generated shapes (two per query).
If a specific CLIP is used in the optimization, then the CLIP R-Precision is more meaningful with a different CLIP encoder (i.e. different backbone).
Papers that use this evaluation protocol include \citet{jain2022zero,lee2022understanding, poole2022dreamfusion,mohammad2022clip,tsalicoglou2023textmesh,xu2023dream3d}.
This metric is an approximation of the \emph{fidelity} of the description to the generated shape.  

\mypara{Shape-text score (ST-S).}
Another metric that measures the \emph{fidelity} of the shape to the text is the shape-text (ST-S) score.
Given an aligned shape-text space, it is possible to compare methods by computing the similarity between the input text and the generated 3D shape.
\citet{zhao2023michelangelo} measured ST-S using ULIP~\cite{xue2023ulip} and their own aligned space SITA~\cite{zhao2023michelangelo}.

\noindent
\mypara{Pretrained LVLM.}
The rise of Large Vision-Language Models (LVLM) is enabling automated metrics that act as a `proxy' for human judgement. \cite{wu2024gpt} demonstrated that GPT-V4 can evaluate the quality of generated 3D shapes through an A2C test for the LVLM.  Given two generated 3D assets, they provide the LVLM with 2D renderings for each, (arranged as a grid of multi-view images), paired with text instructions describing the criteria to judge the renderings (different prompts and different rendering styles are used depending on the criteria of interest).  The LVLM outputs whether the left or right asset is better, together with an analysis. Given a set of text prompts, and a set of models, the pairwise rankings are then combined to form an overall score by using the Elo score (commonly used for chess rankings). This strategy enables automated evaluation of criteria such as the fidelity (or \emph{alignment}) of the text to the 3D shape, \emph{plausibility} of the geometry, \emph{geometry details}, \emph{texture details}, and \emph{diversity} of the shapes.

\section{Discussion}
\label{sec:discussion}

This survey has summarized and categorized work on text-to-3D shape generation.
Despite the explosion of interest in this area, text-to-3D research is still far behind text-to-image generation with many challenges and opportunities.
Below, we conclude by outlining some promising directions for future investigation.

\mypara{Data scale.}
With the development of ever-larger 3D datasets such as Objaverse~\citep{deitke2023objaverse} and Objaverse-XL~\citep{deitke2023objaversexl}, there are increasing amounts of 3D shape data available to the research community.
This trend is likely to continue along with better 3D scanning techniques, and easier 3D content design tools.
This data will will likely be highly valuable for training better generative 3D models.
Moreover, while this data will not necessarily be naturally paired with text descriptions, advances in image captioning will allow generation of text captions at scale.
Thus, we anticipate further opportunities in developing better methods that can leverage both paired and unpaired 3D and text.

\mypara{Hierarchical and part-based generation.}
As was apparent from this survey's section on scene-level generation, much work remains to be done to enable generating hierarchical compositions of scenes from high-quality objects, or similarly, hierarchical compositions of objects from parts.
Progress in this direction will enable fine-grained editability, since many editing operations revolve around properties of specific parts, or specific objects within a scene.
Moreover, generation of animated 3D content whether at the object or scene scale, will likely benefit from such hierarchical methods, as many motions are typically characterized well through rigid transformations of parts or whole objects.
This latter generation of dynamic 3D content is a prominent ``grand challenge'' in this area.

\mypara{Focus on language.}
Much of the development in text-to-3D generation has been driven by advances in 3D representations and generative models from other domains, text to image being the prominent example.
While this transfer of knowledge led to an explosion of interest, an important and currently under-studied area is generation that better matches language to the generated shapes in a fine-grained manner (e.g., respecting detailed part and material properties).
In addition, questions such as what kinds of natural language are handled well by particular methods remain unanswered.
A key challenge for future work is shape generation that respects the fine-grained compositionality of the input language (e.g., "chair with black arms and red seat).
In order to make progress in this direction, the community will also need to devote focus on evaluation protocols that can quantify progress in these directions.

\mypara{Improved speed and memory.}
Methods that rely on 3D data typically require large amounts of compute and memory to train, but are generally fast at inference time.
In comparison, methods that do not rely on 3D data typically rely on per-prompt optimization which is very slow, and currently highly impractical for real-world deployment.
Thus, another promising direction for future work is developing strategies to improve the efficiency, speed, and memory consumption characteristics of text-to-3D shape generation.
At the time of writing this survey, there have been exciting recent developments that leverage 3D Gaussian Splatting~\citep{kerbl20233d} as backbones for the SDS loss~\citep{tang2023dreamgaussian, chen2023text, yi2023gaussiandreamer}, leading to significant improvements in training and rendering speeds.

\mypara{Conclusion.}
We hope that this survey will catalyze further work in text-to-3D shape generation, and enable researchers to advance the state of the art.
Progress in this direction has the potential to democratize 3D content creation by enabling people to turn their imagination into high-quality 3D assets, and to iteratively design and control these assets for a variety of application domains.

\vspace{20pt}
\mypara{Acknowledgments.}
This work was funded in part by a CIFAR AI Chair, a Canada Research Chair and NSERC Discovery Grant.
We thank Ali Mahdavi-Amiri for helpful comments and discussions.

\renewcommand*{\mkbibnamefamily}[1]{\textsc{#1}}
\renewcommand*{\mkbibnamegiven}[1]{\textsc{#1}}
\printbibliography

\end{document}